\documentclass[acmsmall]{acmart}
\usepackage{xcolor}
\usepackage{multirow}
\usepackage{soul}
\usepackage{graphicx}
\usepackage{subcaption}
\usepackage{makecell}

\definecolor{lightpink}{HTML}{ecbdc9}
\definecolor{pastelgreen}{HTML}{c6dec8}
\definecolor{lightyellow}{HTML}{f2e9c0}
\definecolor{paleblue}{HTML}{bfd3e4}
\definecolor{lightpeach}{HTML}{f0d3c4}
\definecolor{lightlavender}{HTML}{d8bedf}
\definecolor{palecyan}{HTML}{c9eeed}

\newcommand{\highlightpink}[1]{\sethlcolor{lightpink}\hl{#1}}

\newcommand{\highlightblue}[1]{\sethlcolor{paleblue}\hl{#1}}
\newcommand{\highlightpeach}[1]{\sethlcolor{lightpeach}\hl{#1}}

\definecolor{darkgreen}{rgb}{0.0, 0.5, 0.0}

\AtBeginDocument{%
  \providecommand\BibTeX{{%
    \normalfont B\kern-0.5em{\scshape i\kern-0.25em b}\kern-0.8em\TeX}}}

\setcopyright{acmlicensed}
\copyrightyear{2025}
\acmYear{2025}
\acmDOI{10.1145/3744744}

\begin{document}

\title{An In-depth Evaluation of Large Language Models in Sentence Simplification with Error-based Human Assessment}

\author{Xuanxin Wu}
\affiliation{%
  \institution{Graduate School of Information Science and Technology, The University of Osaka}
  \streetaddress{1-5 Yamadaoka}
  \city{Suita}
  \state{Osaka}
  \country{Japan}
}
\email{xuanxin.wu@ist.osaka-u.ac.jp}

\author{Yuki Arase}
\affiliation{%
  \institution{School of Computing, Institute of Science Tokyo}
  \streetaddress{2 Chome-12-1 Ookayama}
  \city{Meguro}
  \state{Tokyo}
  \country{Japan}
}
\email{arase@c.titech.ac.jp}

\begin{abstract}
Sentence simplification, which rewrites a sentence to be easier to read and understand, is a promising technique to help people with various reading difficulties. With the rise of advanced large language models (LLMs), evaluating their performance in sentence simplification has become imperative. 
Recent studies have used both automatic metrics and human evaluations to assess the simplification abilities of LLMs. 
However, the suitability of existing evaluation methodologies for LLMs remains in question. 
First, the suitability of current automatic metrics on LLMs' simplification evaluation is still uncertain. 
Second, current human evaluation approaches in sentence simplification often fall into two extremes: they are either too superficial, failing to offer a clear understanding of the models' performance, or overly detailed, making the annotation process complex and prone to inconsistency, which in turn affects the evaluation's reliability. 
To address these problems, this study provides in-depth insights into LLMs' performance while ensuring the reliability of the evaluation. 
% We introduce a novel error-based human evaluation framework and adapt this framework to compare the performance of the most advanced GPT-4 and the state-of-the-art supervised model, Control-T5, to assess their simplification capabilities. 
We design an error-based human annotation framework to assess the LLMs' simplification capabilities. We select both closed-source and open-source LLMs, including GPT-4, Qwen2.5-72B, and Llama-3.2-3B. We believe that these models offer a representative selection across large, medium, and small sizes of LLMs. Results show that LLMs generally generate fewer erroneous simplification outputs compared to the previous state-of-the-art. However, LLMs have their limitations, as seen in GPT-4's and Qwen2.5-72B's struggle with lexical paraphrasing. Furthermore, we conduct meta-evaluations on widely used automatic metrics using our human annotations. We find that these metrics lack sufficient sensitivity to assess the overall high-quality simplifications, particularly those generated by high-performance LLMs\footnote{Our corpus is available at \url{https://github.com/WuXuanxin/human-eval-llm-simplification}}.
\end{abstract}



% \received[accepted]{8 May 2025}
\maketitle
\newpage
\section{Introduction}
Sentence simplification automatically rewrites sentences to make them easier to read and understand by modifying their wording and structures, without changing their meanings. It helps people with reading difficulties, such as non-native speakers~\cite{Gustavo2016Lexical}, individuals with aphasia~\cite{carroll1999simplifying}, dyslexia~\cite{Rello2013DysWebxia, Rello2013Simplify}, or autism~\cite{Barbu2015Language}. 
Previous studies often employed a sequence-to-sequence model, which was then enhanced by integrating various sub-modules into it~\cite{zhang-lapata-2017-sentence, zhao-etal-2018-integrating, nishihara-etal-2019-controllable, martin-etal-2020-controllable}. 
Recent developments have seen the rise of large language models (LLMs). Among them, the closed-source ChatGPT families released by OpenAI demonstrate exceptional general and task-specific abilities~\cite{openai2023gpt4, wang2023Document, liu2023geval}, and sentence simplification is not an exception. On the open-source front, LLMs such as the Llama family by Meta~\cite{touvron2023llamaopenefficientfoundation} and the Qwen family by Alibaba Cloud~\cite{bai2023qwentechnicalreport} stand out as prominent representatives, showing competitive performance.

Some studies~\cite{feng2023sentence, kew-etal-2023-bless} have begun to evaluate LLMs' performance in sentence simplification, including both automatic scoring and conventional human evaluations where annotators assess the levels of fluency, meaning preservation, and simplicity~\cite{kriz-etal-2019-complexity, jiang2020neural, alva-manchego-etal-2021-un, maddela-etal-2021-controllable}, or identify common edit operations~\cite{alva-manchego-etal-2017-learning}.
% \footnote{In the study conducted by Kew et al.~\cite{kew-etal-2023-bless}, while the annotation of common failures are also incorporated, it is noteworthy that the types of failures addressed are very limited and they selected only a handful of output samples for annotation.}. 
However, these studies face limitations and challenges. Firstly, it is unclear whether the current automatic metrics are suitable for evaluating the simplification abilities of LLMs. Although these metrics have demonstrated variable effectiveness across conventional systems (e.g., semantics-informed rule-based~\cite{sulem-etal-2018-simple}, statistical machine translation-based~\cite{wubben-etal-2012-sentence, xu-etal-2016-optimizing}, and sequence-to-sequence model-based simplification~\cite{zhang-lapata-2017-sentence,martin-etal-2020-controllable}) through their correlation with human evaluations \cite{alva-manchego-etal-2021-un}, their suitability for LLMs has yet to be explored, thereby their effectiveness in assessing LLMs' simplifications are uncertain. Secondly, given the general high performance of LLMs, conventional human evaluations may be too superficial to capture the subtle yet critical aspects of simplification quality. This lack of depth undermines the interpretability when evaluating LLMs. Recently, Heineman et al.~\cite{heineman-etal-2023-dancing} proposed a detailed human evaluation framework for LLMs, categorizing $21$ linguistically based success and failure types. However, their linguistics-based approach appears to be excessively intricate and complex, resulting in low consistency among annotators, thus raising concerns about the reliability of the evaluation. The trade-off between interpretability and reliability underscores the necessity for a more balanced approach. 

Our goal is to make a clear understanding of LLMs' performance on sentence simplification, and to reveal whether current automatic metrics are genuinely effective for evaluating LLMs' simplification ability. We design an \textbf{error-based human evaluation framework} to identify key failures in important aspects of sentence simplification, such as inadvertently increasing complexity or altering the original meaning. Our approach aligns closely with human intuition by focusing on outcome-based assessments rather than linguistic details. This straightforward approach makes the annotation easy without necessitating a background in linguistics. Additionally, we conduct a \textbf{meta-evaluation of automatic evaluation metrics} to examine their effectiveness in measuring the simplification abilities of LLMs by utilizing data from human evaluations. 

\begin{figure*}[t]
\centering
\includegraphics[width=0.9\textwidth]{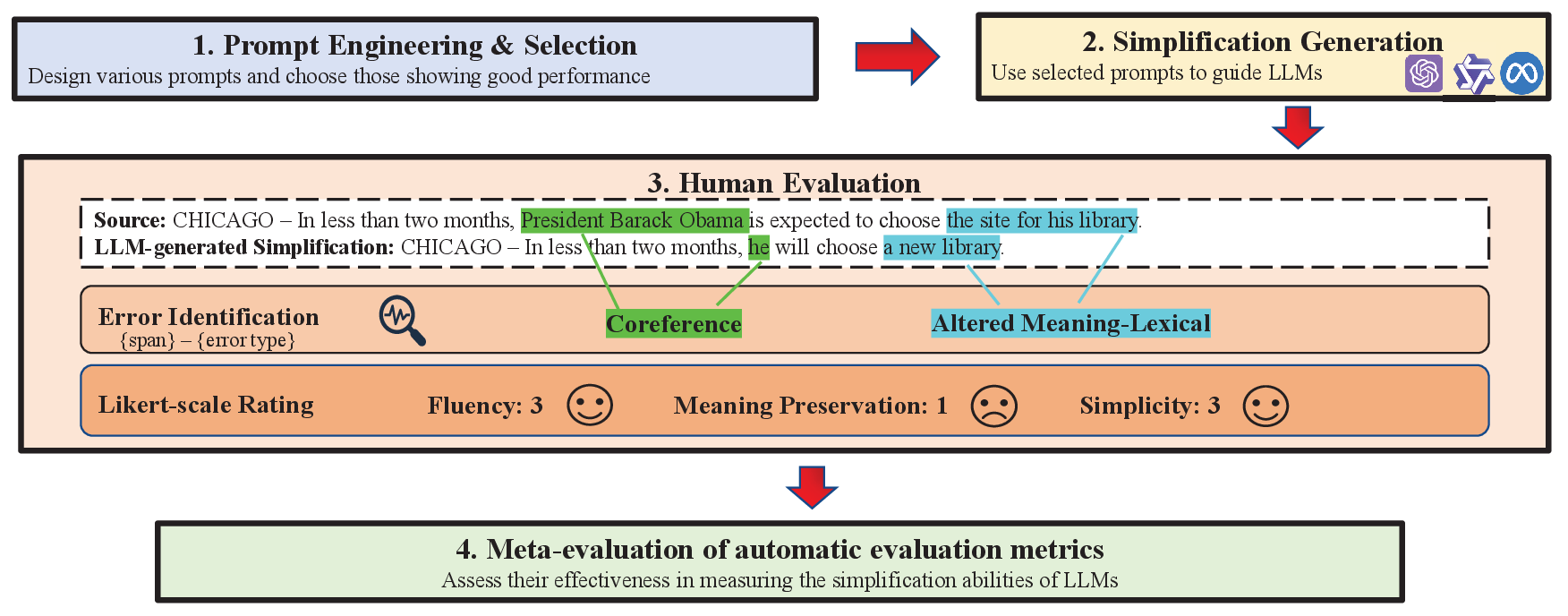}
\caption{Overview of our methodology: sequential evaluation pipeline with human assessment example}
\label{fig:pipeline}
\end{figure*}

We apply our error-based human evaluation framework to evaluate the performance of GPT-4\footnote{We used the `gpt-4-0613' and accessed it via OpenAI's APIs.}, Qwen2.5-72B, and Llama-3.2-3B\footnote{We used the `Qwen2.5-72B-Instruct' and  `Llama-3.2-3B-Instruct'. We ran the two models using Transformers library\cite{HuggingFace_transformer2019}.} in English sentence simplification. We believe that these models offer a representative selection across large, medium, and small sizes of LLMs. We use prompt engineering and evaluate models on four representative datasets on sentence simplification: Turk~\cite{xu-etal-2016-optimizing}, ASSET~\cite{alva-manchego-etal-2020-asset}, Newsela~\cite{xu-etal-2015-problems}, and SimPA~\cite{scarton-etal-2018-simpa}. Figure~\ref{fig:pipeline} illustrates the overview of our evaluation pipeline. Our key findings are summarized as follows:
\begin{itemize}
    \item \textbf{LLMs generally surpass the previous state-of-the-art (SOTA) in performance}; LLMs tend to generate fewer erroneous simplification outputs and better preserve the original meaning, while maintaining comparable levels of fluency and simplicity. 
    \item Among the LLMs, GPT-4 and Qwen2.5-72B surpass Llama-3.2-3B, with Qwen2.5-72B generating fewer errors than GPT-4. This implies the \textbf{strong potential of medium-sized LLMs in simplification tasks}. 
    \item However, larger LLMs have their limitations, as seen in GPT-4 and Qwen2.5-72B’s struggles with \textbf{lexical paraphrasing}.
    \item The meta-evaluation reveals that \textbf{existing automatic metrics struggle to effectively differentiate between high- and low-quality simplifications labeled by human}, particularly when evaluating the overall high-quality outputs of GPT-4 and Qwen2.5-72B.
\end{itemize}

\section{Related Work}
Our study evaluates the performance of representative closed-source and open-source LLMs of varying sizes in sentence simplification by comparing them against the SOTA supervised simplification model. This section includes a review of current evaluations of LLMs in this domain, along with an overview of the SOTA supervised simplification model.

\subsection{Evaluation of LLM-based Simplification}
In sentence simplification, some studies attempted to assess the performance of LLMs. For example, Feng et al.~\cite{feng2023sentence} evaluated the performance of prompting ChatGPT and GPT-3.5; later, Kew et al.~\cite{kew-etal-2023-bless} compared $44$ LLMs varying in size, architecture, pre-training methods, and with or without instruction tuning.  Additionally, Heineman et al.~\cite{heineman-etal-2023-dancing} proposed a detailed human evaluation framework for LLMs, categorizing 21 linguistically based success and failure types. Their findings indicate that OpenAI’s LLMs generally surpass the previous SOTA supervised simplification models.

However, these studies have three primary limitations. First, there has not been a comprehensive exploration into the capabilities of the most advanced closed-source and open-source models to date, i.e., GPT-4 and Llama-3. Second, these studies do not adequately explore prompt variation, employing uniform prompts with few-shot examples across datasets without considering their unique features in simplification strategies. This may underutilize the potential of LLMs, which are known to be prompt-sensitive. Third, the human evaluations conducted are inadequate. Such evaluations are crucial, as automatic metrics often have blind spots and may not always be entirely reliable~\cite{he-etal-2023-blind}. Human evaluations in these studies often rely on shallow ratings or edit operation identifications to evaluate a narrow range of simplification outputs. These methods risk being superficial, overlooking intricate features. In contrast, Heineman et al.’s linguistics-based
approach~\cite{heineman-etal-2023-dancing} appears to be excessively intricate and complex, resulting in low consistency among annotators, thus
raising concerns about the reliability of the evaluations. 
Our study aims to bridge these gaps, significantly enhancing the utility of LLMs through comprehensive prompt engineering processes, and incorporating elaborate human evaluations while ensuring reliability.

\subsection{SOTA Supervised Simplification Models} 
\label{sec:sota}
Traditional NLP methods heavily relied on task-specific models, which involve adapting pre-trained language models for various downstream applications. In sentence simplification, Martin et al.~\cite{martin-etal-2022-muss} introduced the MUSS model by fine-tuning BART~\cite{lewis-etal-2020-bart} with labeled sentence simplification datasets and/or mined paraphrases. Similarly, Sheang et al.~\cite{sheang-saggion-2021-controllable} fine-tuned T5~\cite{Raffel2020t5}, which is called Control-T5 in this study, achieving SOTA performance on two representative datasets: Turk and ASSET. These models leverage control tokens, which were initially introduced by ACCESS~\cite{martin-etal-2020-controllable}, to modulate attributes like length, lexical complexity, and syntactic complexity during simplification. This approach allows any sequence-to-sequence model to adjust these attributes by conditioning on simplification-specific tokens, facilitating strategies that aim to shorten sentences or reduce their lexical and syntactic complexity. Our study employs Control-T5 as the previous SOTA model and compares it to LLMs in sentence simplification.

\section{Datasets}
In this study, we employ standard datasets for English sentence simplification, as detailed below. 
For replicating the SOTA supervised model, namely, Control-T5, we use the same training datasets as the original paper. Meanwhile, the evaluation datasets are used to assess the performance of our models.

\subsection{Training Datasets}
\label{sec:train_dataset}
We use training sets from two datasets: \textbf{WikiLarge}~\cite{zhang-lapata-2017-sentence} and \textbf{Newsela}~\cite{xu-etal-2015-problems,zhang-lapata-2017-sentence}. \textbf{WikiLarge} consists of $296k$ complex-simple sentence pairs automatically extracted from English Wikipedia and Simple English Wikipedia by sentence alignment. Introduced by Xu et al.~\cite{xu-etal-2015-problems}, \textbf{Newsela} originates from a collection of news articles accompanied by simplified versions written by professional editors. It was subsequently aligned from article-level to sentence-level, resulting in approximately $94k$ complex-simple sentence pairs. In our study, we utilize the training split of the Newsela dataset made by Zhang and Lapata~\cite{zhang-lapata-2017-sentence}.

\subsection{Evaluation Datasets}
\label{sec:dataset}
We use validation and test sets from four datasets on English sentence simplification.\footnote{Validation sets from Turk, ASSET, and Newsela were used for prompt engineering on GPT-4.} Table~\ref{tab:datasets} shows the numbers of complex-simple sentence pairs in these sets. These datasets have distinctive features due to differences in simplification strategies and as summarized below.

\begin{itemize}
    \item \textbf{Turk~\cite{xu-etal-2016-optimizing}:} This dataset comprises $2,359$ sentences from English Wikipedia, each paired with eight simplified references written by crowd-workers. It is created primarily focusing on \textbf{lexical paraphrasing}. 
    
    \item \textbf{ASSET~\cite{alva-manchego-etal-2020-asset}:} 
    This dataset uses the same $2,359$ source sentences as the Turk dataset. It differs from Turk by aiming at rewriting sentences with more \textbf{diverse transformations}, i.e., paraphrasing, deleting phrases, and splitting a sentence, and provides $10$ simplified references written by crowd-workers. 

    \item \textbf{Newsela~\cite{xu-etal-2015-problems,zhang-lapata-2017-sentence}:} This is the same Newsela dataset described in Section~\ref{sec:train_dataset}.  
    We utilize its validation and test splits, totaling $2,206$ sentence pairs. After careful observation, we found that \textbf{deletions of words, phrases, and clauses} predominantly characterize the Newsela dataset. 

    \item \textbf{SimPA~\cite{scarton-etal-2018-simpa}:} This dataset originated from the public administration domain. It contains $1,100$ original sentences with two versions of simplified sentences: (1) lexical simplifications (2) lexical and syntactic simplifications. We select the second version for its \textbf{diverse transformations}. Note that SimPA does not provide validation/test splits. We use this dataset exclusively as a test set, excluding three sentence pairs reserved for 3-shot examples.
\end{itemize}

\section{Models}
To enhance the performance of LLMs in sentence simplification, we undertook prompt engineering on GPT-4 across validation datasets, and adapted the prompts for GPT-4, Qwen2.5-72B, and Llama-3.2-3B models. For SimPA, which shares ASSET's diverse transformation characteristic, we reused the optimized instruction for ASSET without additional prompt engineering. We also replicated the SOTA supervised model, Control-T5, for comparative analysis with LLMs. Throughout our optimization efforts, we employed SARI~\cite{xu-etal-2016-optimizing}, which is a widely recognized statistic-based metric for evaluating sentence simplification.\footnote{Our meta-evaluation in Section~\ref{sec:meta_evaluation} confirms that SARI score aligns with human evaluation.} SARI evaluates a simplification model by comparing its outputs against references and source sentences, focusing on the words that are added, kept, and deleted. Its values range from 0 to $100$, with higher values indicating better performance.  

\subsection{LLMs with Prompt Engineering}
By scaling pre-trained language models, such as increasing model and data size, LLMs enhance their capacity for downstream tasks. Unlike earlier models that required fine-tuning, these LLMs can be effectively prompted with zero- or few-shot examples for task-solving. Previous research has looked into different prompting techniques for various tasks. For example, prompt chaining has been studied for summarization~\cite{sun-etal-2024-prompt}. In the area of named entity recognition, the use of special tokens (like @@\#\#) has been found to enhance entity identification~\cite{wang2023gptnernamedentityrecognition}. In our study, we utilize existing human annotation guidelines from these datasets. This method has been employed in earlier studies, demonstrating its effectiveness~\cite{mishra-etal-2022-cross, sainz2024gollie}.

\subsubsection{Design}
Aiming to optimize LLMs' sentence simplification capabilities, we conducted prompt engineering on GPT-4 based on three principal components:
\begin{itemize}
    \item \textbf{Dataset-Specific Instructions:} We tailored instructions to each dataset's unique features and objectives, as detailed in Section~\ref{sec:dataset}. For the Turk and ASSET datasets, we created instructions referring to the guidelines provided to the crowd-workers who composed the references. In the case of Newsela, where such guidelines are unavailable, we created instructions following the styles used for Turk and ASSET, with an emphasis on deletion. Refer to the Appendix~\ref{appendix:prompts} for detailed instructions.
    \item \textbf{Varied Number of Examples:} We varied the number of examples to attach to the instructions: zero, one, and three. 
    \item \textbf{Varied Number of References:} We experimented with a single or multiple (namely, three) simplification references used in the examples. 
    For Turk and ASSET, which are multi-reference datasets, we manually selected one high-quality reference from their multiple references. Newsela, which is basically a single-reference dataset, offers multiple simplification levels for the same source sentences. For this dataset, we extracted references targeting different simplicity levels of the same source sentence as multiple references. 
\end{itemize}
We integrated these components into prompts, resulting in the creation of $15$ variations. These prompts were then applied to each validation set, excluding selected examples. Prompts that achieved the highest SARI scores were designated as `Best Prompts', which are summarized in Table~\ref{tab:prompt-tuning}. For more detailed information, refer to the Appendix~\ref{appendix:prompts}. Following this, we used the best prompts to generate simplification outputs from the respective test sets. 

\subsubsection{Effect of Prompt Engineering}
\label{sec:prompt_engineering}
Prompt engineering demonstrates its effectiveness. As shown in Table~\ref{tab:prompt-tuning}, across three validation sets, prompts with the highest SARI scores significantly outperform those with the lowest, achieving scores of $8.3$ for Turk, $4.5$ for ASSET, and $3.6$ for Newsela. Moreover, results reveal a direct alignment between the best prompt's instructional style and its respective dataset. These top-performing prompts all use a few-shot examples of three. 
The optimal number of simplification references varies; Turk and ASSET show strong results with a single reference, whereas Newsela benefits from multiple references, likely due to the intricacies involved in ensuring that meaning is preserved amidst deletions. Again, SimPA was not included in the prompt engineering process. Instead, we directly applied the instruction from ASSET, accompanied by 3-shot examples with single references from SimPA itself, given the similarity between the two datasets in their emphasis on diverse transformations.
Overall, prompt engineering notably enhances GPT-4's sentence simplification output, as evidenced by the significant increase in SARI.

\begin{table}[t]
\parbox[t]{.35\linewidth}{
\small
\centering
\caption{Number of complex-simple sentence pairs in the validation and test sets of each dataset.}
\label{tab:datasets}
\begin{tabular}{lccl}\toprule
\textbf{Dataset}  & \textbf{Validation} & \textbf{Test} \\ \midrule
Turk                      & $2,000$                 & $359$                      \\ 
ASSET                   & $2,000$                 & $359$                              \\ 
Newsela                  & $1,129$                 & $1,077$                            \\ 
SimPA                  & $0$                 & $1,100$                           \\ \bottomrule
\end{tabular}
}
\parbox{.05\linewidth}{\hfill}
\parbox[t]{.45\linewidth}{
\small
\centering
\caption{The Impact of Prompt Engineering on SARI Scores: Few-Shot (FS), Single Reference (SR), and Multi-Reference (MR)}
\label{tab:prompt-tuning}
\begin{tabular}{lcl}
\toprule
\textbf{Valid Set} & \textbf{SARI Diff.} & \textbf{Best Prompts} \\
\midrule
Turk      & $8.3$ & Turk style + FS + SR \\
ASSET     & $4.5$ & ASSET style + FS + SR \\
Newsela   & $3.6$ & Newsela style + FS + MR \\
\bottomrule
\end{tabular}
}
\end{table}

\subsection{Replicated Control-T5}
\label{sec:replicate_t5}
We replicated the Control-T5 model~\cite{sheang-saggion-2021-controllable}. We started by fine-tuning the T5-base model~\cite{Raffel2020t5} with the WikiLarge dataset and then evaluated it on the ASSET and Turk’s test sets. 
Unlike the original study, which did not train on or evaluate on Newsela, we incorporated this dataset. We employed Optuna~\cite{Akiba2019Optuna} for hyperparameter optimization, a method consistent with the approach used in the original study with the WikiLarge dataset. This optimization process focused on adjusting the batch size, the number of epochs, the learning rate, and the control token ratios. Note that we did not evaluate Control-T5's performance on SimPA since the training dataset is not available. We refer the reader to Appendix~\ref{appendix:t5_config} for the optimal model configuration we achieved.

\section{Human Evaluation}
\label{sec:human_assess}
Automatic metrics provide a fast and cost-effective way for evaluating simplification but struggle to cover all the aspects; they are designed to capture only specific aspects such as the similarity between the output and a reference. Furthermore, the effectiveness of some automatic metrics has been challenged in previous studies~\cite{sulem-etal-2018-bleu, alva-manchego-etal-2021-un}. Human evaluation, which is often viewed as the gold standard evaluation, may be a more reliable method to determine the quality of simplification. 
% Nevertheless, recent studies have underscored human evaluation issues related to reproducibility and reliability in NLP, arising from lack of standardization and under-specification~\cite{belz-etal-2020-disentangling, briakou-etal-2021-review, belz-etal-2023-non}. These issues are evident from reviews of human evaluation experiments on xxx, yyy, zzz published in NLP papers. 
As we discuss in detail in the following section, achieving a balance between interpretability and consistency among annotators is a challenge in sentence simplification. To address this challenge, we have crafted an error-based approach and made efforts in the annotation process, such as mandating discussions among annotators to achieve consensus and implementing strict checks to ensure the quality of assessments.

\subsection{Our approach: Error-based Human Evaluation}
\label{subsec:error-based}
\subsubsection{Challenge in Current Human Evaluation}
Sentence simplification is expected to make the original sentence simpler while maintaining grammatical integrity and not losing important information. A common human assessment approach involves rating sentence-level automatic simplification outputs by comparing them to source sentences in three aspects: fluency, meaning preservation, and simplicity~\cite{kriz-etal-2019-complexity, jiang2020neural, alva-manchego-etal-2021-un, maddela-etal-2021-controllable}. However, sentence simplification involves various transformations, such as paraphrasing, deletion, and splitting, which affect both the lexical and structural aspects of a sentence. Sentence-level scores are difficult to interpret; they do not indicate whether the transformations simplify or complicate the original sentence, maintain or alter the original meaning, or are necessary or unnecessary. Therefore, such evaluation approach falls short in comprehensively assessing the models' capabilities. 

This inadequacy has led to a demand for more detailed and nuanced human assessment methods. Recently, the SALSA framework, introduced by Heineman et al.~\cite{heineman-etal-2023-dancing}, aimed to provide clearer insights through comprehensive human evaluation and consider both the successes and failures of a simplification system. This framework categorizes transformations into $21$ linguistically-grounded
edit types across conceptual, syntactic, and lexical dimensions to facilitate detailed evaluation. However, due to the detailed categorization, it faces challenges in ensuring consistent interpretations across annotators. This inconsistency frequently leads to low inter-annotator agreement, thereby undermining the reliability of the evaluation. We argue that such extensive and fine-grained classifications are difficult for annotators to understand, particularly those without a linguistic background, making it challenging for them to maintain consistency.

\subsubsection{Error-based Human Evaluation}
To overcome the trade-off between interpretability and consistency in evaluations, we design our error-based human evaluation framework. Our approach focuses on identifying and evaluating key failures generated by advanced LLMs in important aspects of sentence simplification. We aim to cover a broad range of potential failures while making the classification easy for annotators. Our approach reduces the categories to seven types while ensuring comprehensive coverage of common failures. In the study on sentence simplification evaluation of LLMs conducted by Kew et al.~\cite{kew-etal-2023-bless}, while the annotation of common failures is also incorporated, it is noteworthy that the types of failures addressed were very limited, and they selected only a handful of output samples for annotation. 

While not intended for LLM-based simplification, a few previous studies have incorporated error analysis to assess their sequence-to-sequence simplification models~\cite{kriz-etal-2019-complexity, cooper-shardlow-2020-combinmt, maddela-etal-2021-controllable}. 
Starting from the error types established in these studies, we included ones that might also be applicable in the outputs of advanced LLMs. 
Specifically, we conducted a preliminary investigation of ChatGPT-3.5 simplification outputs on the ASSET dataset.\footnote{At the time of this investigation, GPT-4 was not publicly available.} 
As a result, we adopted errors of \texttt{Altered Meaning}, issues with \texttt{Coreference}, \texttt{Repetition}, and \texttt{Hallucination}, while omitted errors deemed unlikely, such as the ungrammatical error. 
Additionally, we identified a new category of error based on our investigation: \texttt{Lack of Simplicity}. 
We observed that ChatGPT-3.5 often opted for more complex expressions rather than simpler ones, which is counterproductive for sentence simplification. Recognizing this as a significant issue, we included it in our error types. We also refined the categories for altered meaning and lack of simplicity by looking into the specific types of changes they involve. Instead of listing numerous transformations like the SALSA framework~\cite{heineman-etal-2023-dancing}, we classified these transformations into two simple categories based on their effects on the source sentence: lexical and structural changes. This categorization leads to four error types: \texttt{Lack of Simplicity-Lexical}, \texttt{Lack of Simplicity-Structural}, \texttt{Altered Meaning-Lexical}, and \texttt{Altered Meaning-Structural}.

Table~\ref{tab:definition_errors} summarizes the definition and examples of our target errors. 
Our approach is designed to align closely with human intuition by focusing on outcome-based assessments rather than linguistic details. Annotators evaluate whether the transformation simplifies and keeps the meaning of source components, preserves named entities accurately, and avoids repetition or irrelevant content. This methodology facilitates straightforward classification without necessitating a background in linguistics.

\begin{table}[t]
\small
\centering
\caption{Definitions and Examples of Errors}
\label{tab:definition_errors}
\begin{tabular}{p{2cm} p{1.2cm} p{3cm} p{3cm} p{3cm}}
\toprule
\textbf{Error} & & \textbf{Definition} & \textbf{Source} & \textbf{Simplification} \\
\midrule
	\multirow{9}{*}{Lack of Simplicity} & \multirow{5}{*}{Lexical} & The simplified sentence uses more intricate lexical expression(s) to replace part(s) of the original sentence. & For Rowling, this scene is \textcolor{red}{important} because it \textcolor{red}{shows} Harry's bravery... & Rowling considers the scene \textcolor{red}{significant} because it \textcolor{red}{portrays} Harry's courage... \\
    
	& \multirow{4}{*}{Structural} & The simplified sentence modifies the grammatical structure, and it increases the difficulty of reading. & \textcolor{red}{The other incorporated cities} on the Palos Verdes Peninsula include... & Other cities on the Palos Verdes Peninsula include..., \textcolor{red}{which are also incorporated}. \\
\midrule
	\multirow{8}{*}{Altered Meaning} & \multirow{4}{*}{Lexical} & Significant deviation in the meaning of the original sentence due to lexical substitution(s). & The Britannica was primarily a Scottish \textcolor{red}{enterprise}. & The Britannica was mainly a Scottish \textcolor{red}{endeavor}. \\
	& \multirow{4}{*}{Structural} & Significant deviation in the meaning of the original sentence due to structural changes. & Gimnasia hired \textcolor{red}{first famed Colombian trainer Francisco Maturana, and then Julio César Falcioni}. & Gimnasia hired \textcolor{red}{two famous Colombian trainers, Francisco Maturana and Julio César Falcioni}. \\
\midrule
\multirow{5}{*}{Coreference} &  &A named entity critical to understanding the main idea is replaced with a pronoun or a vague description. & \textcolor{red}{Sea slugs dubbed sacoglossans} are some of the most... & \textcolor{red}{These} are some of the most... \\
\midrule
\multirow{4}{*}{Repetition} &  &Unnecessary duplication of sentence fragments & The report emphasizes the \textcolor{red}{importance} of sustainable practices. & The report emphasizes the \textcolor{red}{importance, the significance, and the necessity} of sustainable practices. \\
\midrule
\multirow{6}{*}{Hallucination} &  &Inclusion of incorrect or unrelated information not present in the original sentence. & In a short video promoting the charity Equality Now, Joss Whedon confirmed that "Fray is not done, Fray is coming back. & Joss Whedon confirmed in a short promotional video for the charity Equality Now that Fray will return, \textcolor{red}{although the story is not yet finished}. \\   
\bottomrule
\end{tabular}
\end{table}

\subsection{Annotation Process}
\label{subsec:annotation_process}
\begin{figure*}[t]
\centering
\includegraphics[width=\textwidth]{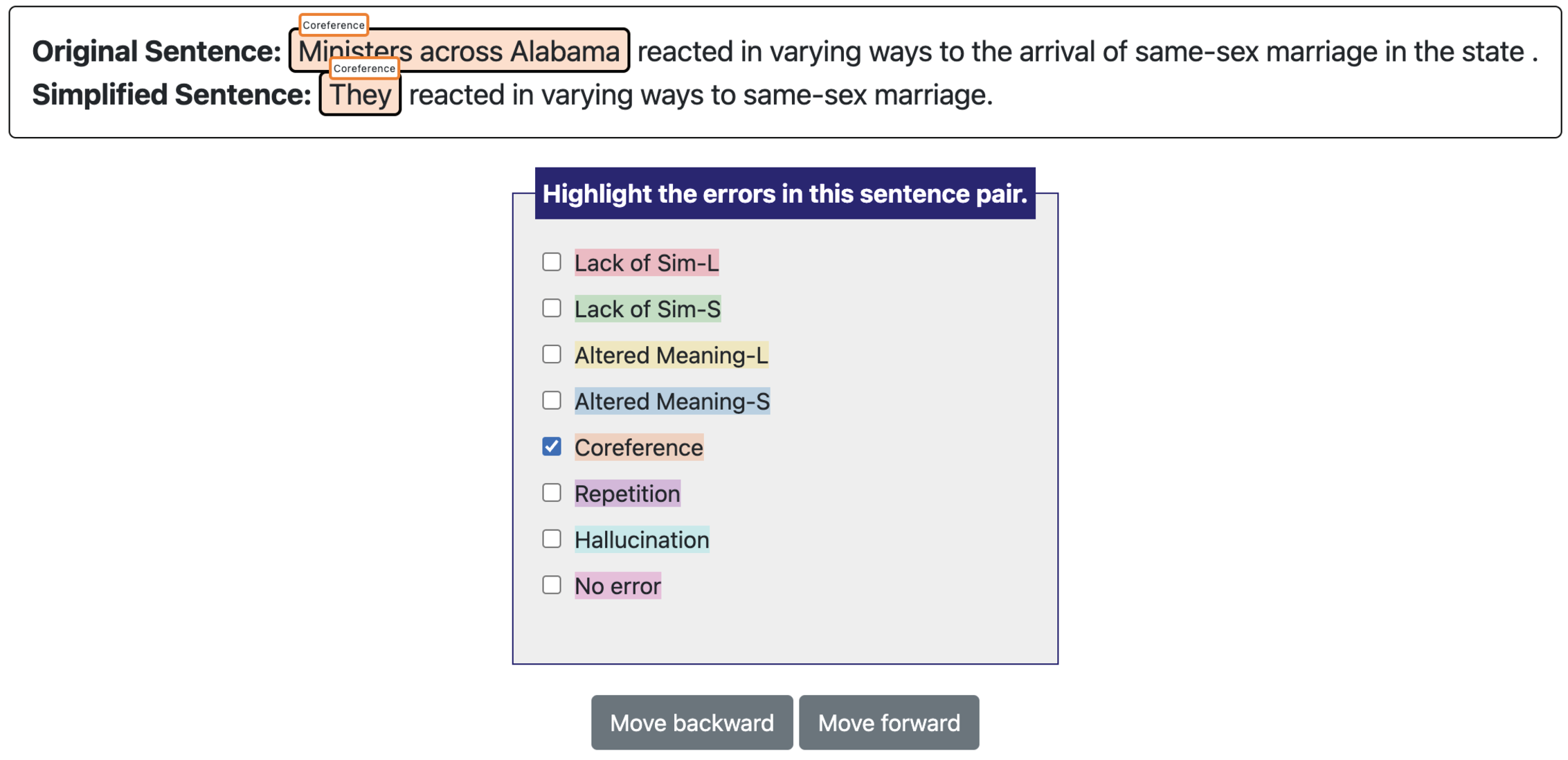}
\caption{Annotation interface in error-based human assessment}
\label{fig:error_interface}
\end{figure*}

We implemented our error-based human evaluation alongside the common evaluation on fluency, meaning preservation, and simplicity using a 1-3 Likert scale. The Potato Platform~\cite{pei2022potato} was utilized to establish our annotation environment for the execution of both tasks. The annotation interface for Task $1$ is illustrated in Figure~\ref{fig:error_interface}. 
Annotators select the error type, marking erroneous spans in the simplified sentence and, when applicable, the corresponding spans in the original (source) sentence. Note that the spans of different error types can overlap each other. 

Each annotator received individual training through a $2.5$-hour tutorial, which covered guidelines and instructions on how to use the annotation platform. Our error-based human evaluations include guidelines that define and provide examples of each error type, as outlined in Table~\ref{tab:definition_errors}. Additionally, detailed guidelines for Likert scale evaluation can be found in the Appendix~\ref{appendix:annotation}.

\subsubsection{Task $1$: Error Identification}
Task $1$ follows our error-based human evaluation detailed in Section~\ref{subsec:error-based}. We sampled $300$ source sentences from Turk, ASSET, and Newsela test sets, along with simplification outputs generated by GPT-4, Qwen2.5-72B, Llama-3.2-3B, and Control-T5, resulting in a total of $3,600$ complex-simple sentence pairs. Additionally, we sampled $300$ source sentences from SimPA and generated simplification outputs using the three LLMs, contributing an additional $900$ complex-simple sentence pairs. Altogether, this resulted in $4500$ complex-simple sentence pairs. The annotation process was conducted over two periods: from October $2023$ to February $2024$, and from October $2024$ to December $2024$.

Annotators were instructed to identify and label errors within each sentence pair according to predefined guidelines. To overcome the trade-off between detailed granularity and annotator agreement, all annotators involved in this task participated in discussion sessions led by one of the authors. These sessions required annotators to share their individual labelings, which were then collectively reviewed during discussions until a consensus was reached. There were $20$ discussion sessions, each lasting approximately three hours, for a total of $60$ hours.

\paragraph{Annotator Selection and Compensation for Task $1$}
As annotators, we used second-language learners with advanced English proficiency, expecting that they would be more sensitive to the fine-grained level of variations in textual difficulty based on their language-learning experiences. 
In addition, given that second language learners stand to benefit significantly from sentence simplification applications, involving them as evaluators seems most appropriate. 
All of our annotators were graduate students or alumni associated with our organization. The compensation rate for this task was set at ¥$100$ JPY (approximately \$$0.67$ USD) per sentence pair. For quality control, annotators had to pass a qualification test before participating in the task. This qualification test comprises annotation guidelines and four complex-simple sentence pairs. Each pair contains various errors predefined by the author. All submissions to this test were manually reviewed. Seven annotators were selected for this task based on their high accuracy in identifying errors, including specifying the error type, location, and rationale. They come from Brazil, China, Italy, Indonesia, and Israel.

\subsubsection{Task $2$: Likert Scale Rating}
Following the convention of previous studies, we also include the rating approach on fluency, meaning preservation, and simplicity using a 1 to 3 Likert scale as Task $2$. In this task, annotators evaluate all simplification outputs generated by LLMs and Control-T5 across four test sets, by comparing them with their corresponding source sentences. In particular, for the Newsela dataset, reference simplifications from the test set were also included. We assume that models trained or tuned on this dataset, which is characterized by deletion, may produce shorter outputs, potentially impacting meaning preservation scores. To ensure fairness, we compare the human evaluation of model-generated simplifications against that of Newsela reference simplifications for a more objective evaluation. The evaluation in Task $2$ covered a total of 11,548 complex-simple sentence pairs. The annotation process was conducted over two periods: from October $2023$ to February $2024$, and from October $2024$ to December $2024$.

To address the challenge of annotator consistency, we implemented specific guidelines during the annotation phase. Annotators were advised to avoid neutral positions (`2' on our scale) unless faced with genuinely challenging decisions. This approach encouraged a tendency towards binary choices, i.e., `1' for simplification outputs that are disfluent, lose a lot of original meaning, or are not simpler, and `3' for simplification outputs that are fluent, preserve meaning, and are much simpler. To ensure quality, one of the authors reviewed $200$ pairs of sampled submissions from each annotator. If any issues were identified, such as an annotator rating inconsistently for sentence pairs with similar problems, they were required to revise and resubmit their annotations.

\paragraph{Annotator Selection and Compensation for Task $2$}
Same with Task $1$, we used second-language learners with advanced English proficiency, who were graduate students or alumni associated with our organization. 
Additionally, we included a native speaker with English education experience, as this candidate's evaluations demonstrated comparable reliability in overall assessment quality. 
Annotator candidates had to pass a qualification test before participating in the task. This qualification test comprises annotation guidelines and five complex-simple sentence pairs. Candidates were instructed to rate fluency, meaning preservation, and simplicity on each simplification output. Seven annotators were selected based on their high inter-annotator agreement, demonstrated by the Intraclass Correlation Coefficient (ICC)~\cite{shrout1979intraclass} score of $0.62$, indicating a substantial agreement. They come from Brazil, China, Italy, Indonesia, Malaysia, and the United States. The compensation rate for this task was set at ¥$40$ JPY (approximately \$$0.27$ USD) per sentence pair. 

\paragraph{Inter-Annotator Agreement}
We assess inner-annotator agreement through the overlapping rate of ratings across three annotators, as detailed in Table~\ref{tab:overlap}. The overlapping rate is calculated by the proportion of identical ratings for a given simplification output.\footnote{We also tried Fleiss' kappa, Krippendorff's alpha, and ICC; however, they resulted in degenerate scores due to too-high agreements on mostly binary judgments.} In the fluency dimension, all models demonstrate strong agreement, with overlapping rates between $96.9\%$ and $99.7\%$. In meaning preservation and simplicity, these dimensions exhibit comparably more variability in ratings, with a broader range of agreement. Annotators found it more subjective to assess meaning preservation and simplicity, as these aspects required direct comparison with the source sentences. Nevertheless, mid to high agreement levels are still achieved, showing the consistency of our annotation.

% \begin{table}[t]
% \small
% \centering
% \caption{Overlapping rate across three annotators (\%)}
% \label{tab:overlap}
% \begin{tabular}{lccccccc}
% \toprule
% & \multicolumn{2}{c}{Turk} & \multicolumn{2}{c}{ASSET} & \multicolumn{3}{c}{Newsela} \\
% \cmidrule(lr){2-3} \cmidrule(lr){4-5} \cmidrule(lr){6-8}
% Dimension & GPT-4 & T5 & GPT-4 & T5 & GPT-4 & T5 & Reference \\
% \midrule
% Fluency    & $98.1$ & $97.8$ & $97.8$ & $96.9$ & $99.6$ & $98.4$ & $98.0$ \\
% Meaning Preservation    & $86.4$ & $57.9$ & $84.7$ & $58.2$ & $66.6$ & $87.3$ & $62.6$ \\
% Simplicity & $77.7$ & $85.5$ & $68.2$ & $90.3$ & $81.3$ & $95.5$ & $76.5$ \\
% % \midrule
% % \textbf{Avg} & $87.4$ & $80.4$ & $83.6$ & $81.8$ & $82.5$ & $93.7$ & $79.0$ \\
% \bottomrule
% \end{tabular}
% \end{table}

\begin{table}[t]
\small
\centering
\caption{Overlapping rate across three annotators (\%)}
\label{tab:overlap}
\resizebox{\textwidth}{!}{%
\begin{tabular}{lcccccccccccccccc}
\toprule
& \multicolumn{4}{c}{Turk} & \multicolumn{4}{c}{ASSET} & \multicolumn{5}{c}{Newsela} & \multicolumn{3}{c}{SimPA} \\
\cmidrule(lr){2-5} \cmidrule(lr){6-9} \cmidrule(lr){10-14} \cmidrule(lr){15-17}
Dimension & GPT-4 & Qwen & Llama & T5 & GPT-4 & Qwen & Llama & T5 & GPT-4 & Qwen & Llama & T5 & Ref. & GPT-4 & Qwen & Llama \\
\midrule
Fluency & $98.1$ & $98.9$ & $99.2$ & $97.8$ & $97.8$ & $99.7$ & $98.6$ & $96.9$ & $99.6$ & $99.4$ & $99.4$ & $98.4$ & $98.0$ & $98.4$ & $98.2$ & $99.2$ \\
Meaning & $86.4$ & $87.2$ & $58.8$ & $57.9$ & $84.7$ & $88.3$ & $70.8$ & $58.2$ & $66.6$ & $59.4$ & $55.5$ & $87.3$ & $62.6$ & $83.7$ & $85.4$ & $54.3$ \\
Simplicity & $77.7$ & $90.5$ & $82.2$ & $85.5$ & $68.2$ & $83.0$ & $64.1$ & $90.3$ & $81.3$ & $91.2$ & $82.4$ & $95.5$ & $76.5$ & $88.5$ & $91.7$ & $90.2$\\
\bottomrule
\end{tabular}%
}
\end{table}

\section{Annotation Result Analysis}
Our comprehensive analysis of annotation data reveals that, overall, LLMs generate fewer erroneous simplification outputs compared to Control-T5, demonstrating higher ability in simplification. Larger models, such as GPT-4 and Qwen2.5-72B, excel at preserving meaning compared to smaller models like Llama-3.2-3B and Control-T5. However, larger LLMs are not without flaws; their most common error is replacing simpler lexical expressions with more complex ones.

\subsection{Analysis of Task 1: Error Identification}
\begin{figure*}[t]
\centering
\includegraphics[width=0.8\textwidth]{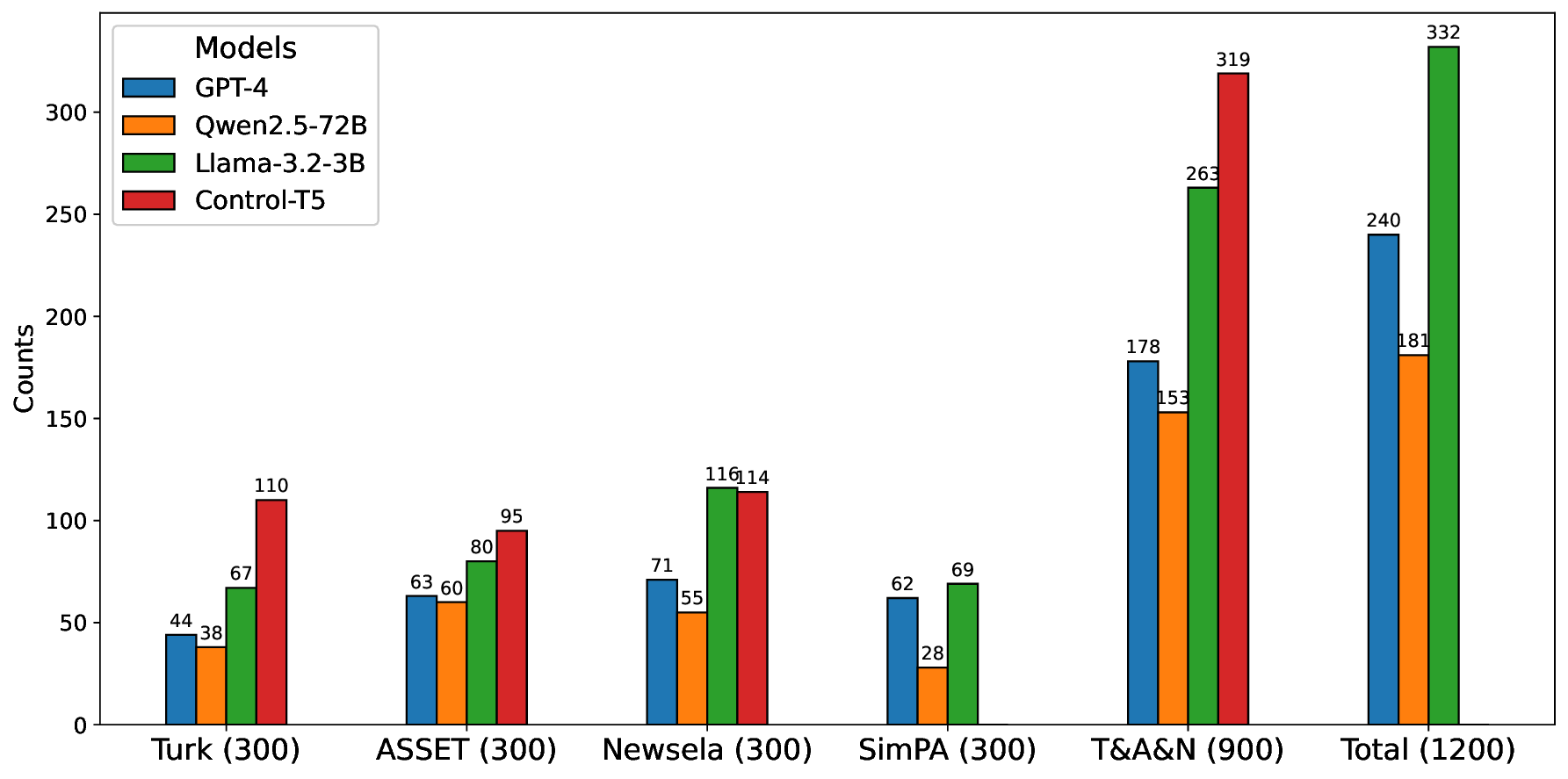}
\caption{Comparison of the number of erroneous simplification outputs generated by models. Numbers in the square brackets represent the number of samples in datasets.}
\label{fig:error_count_sents}
\end{figure*}

This section presents a comparative analysis of erroneous simplification outputs generated by GPT-4, Qwen2.5-72B, Llama-3.2-3B, and Control-T5, focusing on error quantification and type analysis.
We assess erroneous simplification outputs across four datasets: Turk, ASSET, Newsela, and SimPA, defining an erroneous output as one containing at least one error. As mentioned in Section~\ref{sec:replicate_t5}, we did not include SimPA for Control-T5. To reduce bias stemming from this particular dataset, we also report the results after excluding SimPA, i.e., only on Turk, ASSET, and Newsela (denoted as `T\&A\&N'). Figure~\ref{fig:error_count_sents} shows that the three LLMs generally with fewer errors than Control-T5. This performance difference underscores LLMs' superior performance in simplification tasks. Within the LLM group, Qwen2.5-72B produced the fewest errors, followed by GPT-4, while Llama-3.2-3B generated the most.

% \begin{table}[t]
%   \centering
%   \caption{Comparison of the number of erroneous simplification outputs generated by models.}
%   \label{tab:num_erroneous_simplifications}
%   \begin{tabular}{lcccc}
%     \toprule
%     Test set       & GPT-4 & Qwen2.5-72B & Llama-3.2-3B & Control-T5 \\
%     \midrule
%     Turk ($300$ samples)    & $44$    & $38$          & $67$          & $110$        \\
%     ASSET ($300$ samples)   & $63$    & $60$          & $80$          & $95$        \\
%     Newsela ($300$ samples) & $71$    & $55$          & $116$          & $114$        \\ 
%     SimPA ($300$ samples)   & $62$     & $28$          & $69$          & ---          \\ \hline
%     \textbf{T\&A\&N} ($900$ samples) & $178$    & $154$         & $263$         & $319$        \\
%     \textbf{Total} ($1200$ samples)  & $240$    & $182$         & $332$         & ---          \\ 
%     \bottomrule
%   \end{tabular}
% \end{table}

\subsubsection{Error Co-occurrence}
\label{sec:error_co}
Multiple types of errors may co-occur in a simplification output. Consider the following example where simplification is generated by the Control-T5 model:
\begin{description}
    \item[Source:] CHICAGO -- In less than two months, \highlightpeach{President Barack Obama} is expected to choose \highlightblue{the site for his library}.
    \item[Control-T5:] CHICAGO -- In less than two months, \highlightpeach{he} will choose \highlightblue{a new library}.
\end{description}
In this example, \texttt{Coreference} and \texttt{Altered Meaning-Lexical} errors co-occur. The simplification process replaces \highlightpeach{President Barack Obama} with \highlightpeach{he}, leading to a coreference error. Additionally, the phrase \highlightblue{the site for his library} is oversimplified to \highlightblue{a new library}, thus altering the original meaning. We find that on average, GPT-4-generated erroneous simplification outputs contain $1.08 \pm \scriptstyle{0.28}$ unique errors,
Qwen2.5-72B-generated ones contain $1.04 \pm \scriptstyle{0.22}$, Llama-3.2-3B-generated ones contain $1.14 \pm \scriptstyle{0.4}$, and Control-T5-generated ones contain $1.05 \pm \scriptstyle{0.24}$. These averages are calculated by dividing the sum of unique errors in each erroneous simplification output by the total number of these simplification outputs. This indicates that erroneous simplification outputs across all models typically include only one type of error.

\subsubsection{Distribution of Same Errors}
Section~\ref{sec:error_co} indicates that an erroneous simplification output contains a unique error type on average, then what is the distribution of the same error types in those simplification outputs? 
The same type of error can occur multiple times within a single simplification output. Consider the following example where the simplification output is generated by GPT-4:
\begin{description}
    \item[Source:] In 1990, she was the \highlightpink{only} female entertainer \highlightpink{allowed} to perform in Saudi Arabia.
    \item[GPT-4:] In 1990, she was the \highlightpink{sole} woman performer \highlightpink{permitted} in Saudi Arabia.
\end{description}
In this example, \texttt{Lack of Simplicity-Lexical} error appears twice. The simplification uses more difficult words, \highlightpink{sole} and \highlightpink{permitted}, to replace \highlightpink{only} and \highlightpink{allowed}, respectively.  
% We plotted the label-wise distribution for all models, as illustrated in Figure~\ref{fig:error_labelwise} for Qwen2.5-72B and Control-T5. Results for GPT-4 and Llama-3.2-3B are provided in Figure~\ref{fig:gpt4_llama_labelwise} in Appendix~\ref{appendix:distribution}.
Upon a close observation, we find that across all models, each type of error occurs once in most of the erroneous simplification outputs, and only a small fraction of simplification outputs exhibit the same error type more than once. The maximum repetition of the same error type is capped at four.

\subsubsection{Characteristic Errors in Models}
\label{sec:errors_models}
\begin{figure*}[t]
\centering
\includegraphics[width=0.8\textwidth]{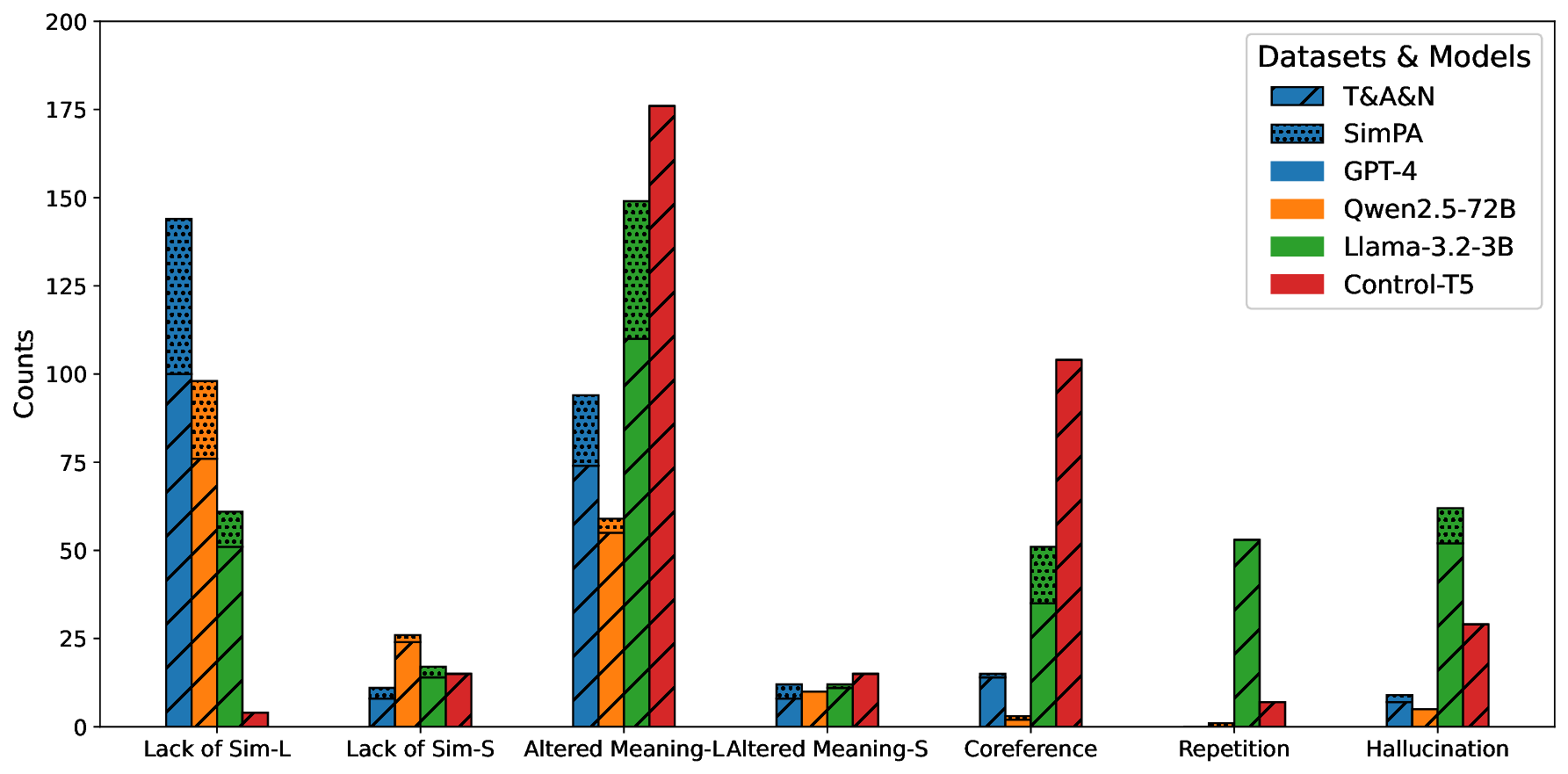}
\caption{Error type distribution across models: each bar represents the count of a specific error type for each model. To ensure a fair comparison between LLMs and Control-T5, we use  `/' and `.' in the graph to distinguish results before (`T\&A\&N') and after incorporating SimpA.}
\label{fig:error_type}
\end{figure*}

We quantitatively analyze the frequency of different error types in simplification outputs generated by models. The results, shown in Figure~\ref{fig:error_type}, indicate variations in error tendencies across models.

\paragraph{LLMs Outperform Control-T5} Consistent with the sentence-level results in Figure~\ref{fig:error_count_sents},
% Table~\ref{tab:num_erroneous_simplifications}, 
\textbf{Control-T5} generates more errors overall ($350$ occurrences) than the LLM group ($211$ for GPT-4, $172$ for Qwen2.5-72B, and $326$ for Llama-3.2-3B after excluding SimPA). Among the LLMs, \textbf{Qwen2.5-72B} produces the fewest errors, followed by GPT-4, with Llama-3.2-3B generating significantly more errors ($202$ for Qwen2.5-72B, $285$ for GPT-4, and $405$ for Llama-3.2-3B). Notably, Qwen2.5-72B performs best in four out of seven error categories, suggesting that while larger LLMs generally perform better, performance may not always scale directly with model size in simplification.

\paragraph{Lexical Paraphrasing is the Biggest Challenge} 
Both \textbf{GPT-4} and \textbf{Qwen2.5-72B} show similar tendencies, with errors predominantly from \texttt{Lack of Simplicity-Lexical} ($144$ for GPT-4 and $98$ for Qwen2.5-72B) and \texttt{Altered Meaning-Lexical} ($94$ for GPT-4 and $59$ for Qwen2.5-72B). This reflects their propensity to employ complex lexical expressions or misinterpret meanings through lexical choices, though Qwen2.5-72B performs better in these categories. \textbf{Control-T5} shows notably high frequencies in \texttt{Altered Meaning-Lexical} ($176$ occurrences) and \texttt{Coreference} ($104$ occurrences). This indicates difficulties with preserving original lexical meanings and ensuring referential clarity. 
Across \textbf{all models}, errors in lexical aspect (\texttt{Lack of Simplicity-Lexical}, \texttt{Altered Meaning-Lexical}, \texttt{Coreference}, \texttt{Repetition}) surpass the occurrences of errors in structural aspect (\texttt{Lack of Simplicity-Structural}, \texttt{Altered Meaning-Structural}) as a general tendency.

\paragraph{Newsela Poses Coreference Resolution Challenge}
Further analysis of Newsela reveals dataset-specific challenges. Compared to Turk and ASSET, \textbf{Control-T5} generates significantly more \texttt{Coreference} errors ($7$ for Turk, $1$ for ASSET, and $96$ for Newsela). After a manual inspection, we find that a possible reason is the high occurrence of coreference within this dataset, and Control-T5 tends to overfit during fine-tuning. 

\paragraph{{Llama-3 is Prone to Repetition Error}}
Remarkably, for \textbf{Llama-3.2-3B}, while paraphrasing remains a significant issue, errors such as \texttt{Repetition} and \texttt{Hallucination} are notably more frequent than in other models. For repetition, some of the errors may stem from the model's misunderstanding of the prompt crafted for Newsela, obtaining from prompt engineering on GPT-4 (see Section~\ref{sec:prompt_engineering}). That is, providing $3$-shot examples with multiple simplification references under each example. Llama-3.2-3B appears to combine multiple simplifications into a single output, leading to repetitive content. Below is an example:
\begin{description}
    \item[Source:] But landowner Gene Pfeifer refused to give up his 3-acre riverfront property in the middle of the proposed library site.
    \item[Llama:] Gene Pfeifer didn't want to sell his 3-acre land.    Gene Pfeifer refused to sell his land.    Gene Pfeifer didn't want to give up his 3-acre property.
\end{description}

\subsection{Likert Scale Rating}
\label{sec:likert_scale_rating}
In this section, we compare model performances across various dimensions and datasets by averaging annotators' ratings. Results show that, as a general tendency, LLMs again consistently outperform Control-T5 across all datasets, indicating a preference among annotators for the LLMs' simplification quality. Among the LLM group, GPT-4 and Qwen2.5-72B show comparable performance, consistently rated higher than Llama-3.2-3B across all datasets.

\begin{figure*}[t]
\centering
\includegraphics[width=0.6\textwidth]{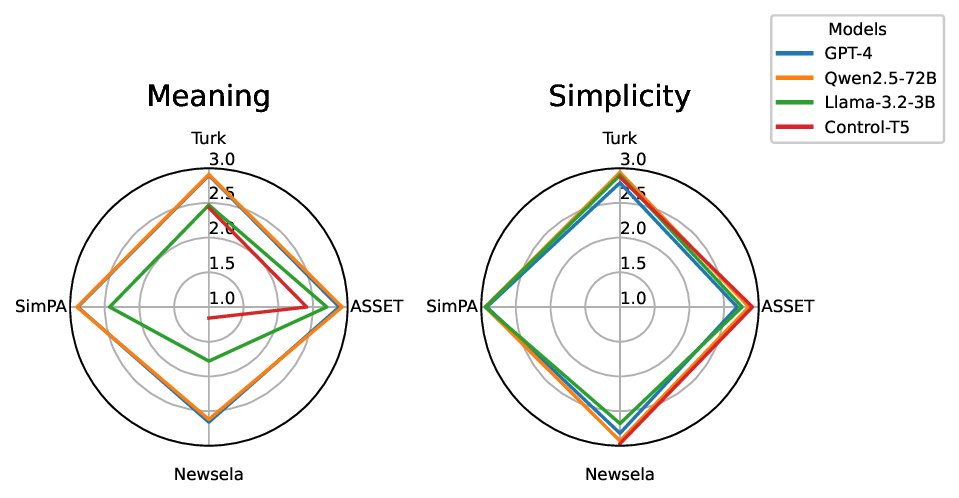}
\caption{Comparison of meaning Preservation and simplicity across models and datasets}
\label{fig:likert_radar}
\end{figure*}

For \textbf{fluency}, all models demonstrate high fluency levels, indicated by the average ratings approaching three. This suggests that these models generate grammatically correct simplifications without significant differences in fluency.
In terms of \textbf{meaning preservation}, GPT-4 and Qwen2.5-72B outperform Llama-3.2-3B and Control-T5 across all datasets: achieving scores of ($2.90$ (GPT-4), $2.91$ (Qwen)) versus ($2.47$ (Llama), $2.43$ (T5)) on Turk, ($2.88$, $2.92$) versus ($2.70$, $2.41$) on ASSET, ($2.66$, $2.62$) versus ($1.78$, $1.16$) on Newsela, and ($2.89$, $2.90$) versus ($2.43$) on SimPA. Conversely, for \textbf{simplicity}, GPT-4 and Qwen2.5-72B's ratings are comparable or slightly lower than those of Llama-3.2-3B and Control-T5, though the disparities are less pronounced than those observed for meaning preservation: ($2.79$ (GPT-4), $2.94$ (Qwen)) versus ($2.90$ (Llama), $2.87$ (T5)) on Turk, ($2.69$, $2.86$) versus ($2.76$, $2.91$) on ASSET, ($2.82$, $2.93$) versus ($2.68$, $2.97$) on Newsela, and ($2.92$, $2.94$) versus ($2.94$) on SimPA. To facilitate a clearer comparison, we provide Figure~\ref{fig:likert_radar}, which visually contrasts the models’ performance in meaning preservation and simplicity. The contrast suggests that Llama-3.2-3B and Control-T5 may be comparably good or slightly better at generating simpler outputs but at the cost of losing a significant portion of the original meaning. 

On the Newsela dataset, we observe that Control-T5 often simply deletes long segments of the source sentence, leaving only trivial changes to the remaining parts. Below is an example:

\begin{description}
    \item[Source:] Sun said his fascination with the past started with his grandfather, who taught him the old accents and ways of Beijing, including Beijing opera.
    \item[Control-T5:] Sun said his fascination with the past started with his grandfather.
    \item[GPT-4:] Sun's grandfather taught him about the culture of old Beijing, which sparked his interest in the past.
    \item[Qwen:] Sun's interest in the past began with his grandfather, who taught him about old Beijing, including its opera.
    \item[Llama:] Sun's interest in the past started with his grandfather, who taught him about Beijing's old ways.
\end{description}

In this example, the source sentence consists of a main clause and a relative clause. Control-T5 removes the relative clause entirely while leaving the main clause unchanged, leading to significant information loss and limited transformation diversity. In contrast, GPT-4, Qwen2.5-72B, and Llama-3.2-3B tend to retain information from both parts, achieving higher scores for meaning preservation.
The average rating for reference simplifications in the Newsela test set is $1.67$, which reveals that Newsela's references sacrifice meaning preservation for simplicity. Control-T5 might have adopted a deletion-heavy approach during its training on the Newsela dataset, even heavier than the deletion degree of the Newsela dataset itself. It adversely affects its ability to preserve the original sentence's meaning.

\subsection{Discussion: Additional Observations}
During the annotation process, we observed several nuanced phenomena that were difficult to fit into specific error categories. Some cases fail to meet the criteria for satisfactory simplifications, including redundancy, lack of common sense, and change of focus. 
Other cases could be controversial, including the addition of factual information not inferable from the source sentence. This section outlines the models where these phenomena were observed and provides the number of reported cases during error identification annotation, along with examples for each category.

\subsubsection{Redundancy}
Simplifications introduced redundancies, failing to contribute meaningfully to the simplification of the sentence or to enhance clarity. This was only observed in Control-T5's simplifications with four reported cases. In the example below, Control-T5 replaces `biochemist' with `biochemist and scientist, ' which seems redundant due to the overlapping parts in meanings.

\begin{description}
    \item[Source:] Their granddaughter Hélène Langevin-Joliot is a professor of nuclear physics at the University of Paris, and their grandson Pierre Joliot, who was named after Pierre Curie, is a noted \underline{biochemist}.
    \item[Control-T5:] Their granddaughter Hélène Langevin-Joliot is a professor of nuclear physics at the University of Paris. Their grandson Pierre Joliot is also a well-known \underline{biochemist and scientist}.
\end{description}

\subsubsection{Lack of Common Sense}
Simplifications that result in logical inconsistencies or nonsensical interpretations. This was only observed in Control-T5's simplifications with three reported cases. In the example below, Control-T5 illogically suggests that Orton gave birth to his wife, indicating a lack of common sense.

\begin{description}
    \item[Source:] Orton and his wife \underline{welcomed} Alanna Marie Orton on July 12, 2008.
    \item[Control-T5:] Orton \underline{gave birth to his wife}, Alanna Marie, on July 12, 2008.
\end{description}

\subsubsection{Change of Focus}
Simplifications that inappropriately alter the original sentence's focus, leading to misleading interpretations. This was only observed in Control-T5 with four reported cases and Llama-3.2-3B with six. In the first example below, Control-T5 shifts the focus from the type of piece Opus 57 to the mere fact that Chopin composed it. Similarly, in the second example, Llama-3.2-3B redirects attention from the agreement and actions of other judges to the federal court's decision itself.
\begin{description}
    \item[Source:] Frédéric Chopin's Opus 57 is a berceuse for solo piano.
    \item[Control-T5:] Frédéric Chopin wrote a piece called Opus 57 for solo piano.
\end{description}
\begin{description}
    \item[Source:] Other judges agreed with the federal court's decision and started marrying same-sex couples in the morning.
    \item[Llama:] The federal court ruled that same-sex couples could get married.
\end{description}

\subsubsection{Factual Information Not Inferable from the Source Sentence}
We found cases where information not explicitly present in the source sentence was added to the simplifications. This was observed in all models, with four reported cases in GPT-4, $12$ in Qwen2.5-72B, five in Llama-3.2-3B, and $12$ in Control-T5. These additions are generally factual and, although not inferable from the source sentence, were verified to be factual using online sources. This type of information can be controversial as it does not strictly adhere to the input. However, it may facilitate the reader's understanding of the source sentence. We did not classify these cases as errors but instead documented them along with external resources, such as website links, to verify their accuracy. For example, in the case below, ``Lincoln's assassination'' cannot be inferred directly from the source sentence. However, Qwen2.5-72B includes this detail, likely drawing on its internal knowledge by linking the provided date and named entities. In such situations, we verified the added information online. If the information was factual, we did not classify it as an error.
\begin{itemize}
    \item \textbf{Source:} For example, there's a letter of sympathy from Queen Victoria to Mary Todd Lincoln on April 29, 1865, calling \underline{his assassination} ``so terrible a calamity''.
    \item \textbf{Qwen:} Queen Victoria wrote a letter of sympathy to Mary Todd Lincoln about \underline{Lincoln's} \underline{assassination}.
\end{itemize}

\section{Meta-Evaluation of Automatic Evaluation Metrics}
\label{sec:meta_evaluation}
Due to the high cost and time requirements of human evaluation, automatic metrics are preferred as a means of obtaining faster and cheaper evaluation of simplification models. Previous studies have explored the extent of widely-used metrics in sentence simplification can assess the quality of outputs generated by neural systems~\cite{sulem-etal-2018-bleu, alva-manchego-etal-2021-un, tanprasert-kauchak-2021-flesch}. However, it remains uncertain whether these metrics are adequately sensitive and robust to differentiate the quality of simplification outputs generated by advanced LLMs, i.e., GPT-4, especially given the generally high performance. To fill this gap, we perform a meta-evaluation of commonly used automatic metrics at both sentence and corpus levels, utilizing our human evaluation data.

\subsection{Automatic Metrics}
\label{sec:metrics}
In this section, we review evaluation metrics that have been widely used in sentence simplification, categorizing them based on their primary evaluation units into two types: sentence-level metrics, which evaluate individual sentences, and corpus-level metrics, which assess the system-wise quality of simplification outputs.

\subsubsection{Sentence-level Metrics}
\begin{itemize}
    \item \textbf{LENS}~\cite{maddela-etal-2023-lens} is a model-based evaluation metric that leverages RoBERTa~\cite{liu2019roberta} trained to predict human judgment scores, considering both the semantic similarity and the edits comparing the output to the source and reference sentences. Its values range from $0$ to $100$, where higher scores indicate better simplifications.
    \item \textbf{BERTScore}~\cite{zhang2020bertscore} provides similarity scores (precision, recall, and f1) for each token in the candidate sentence against each token in the reference, leveraging BERT's~\cite{devlin-etal-2019-bert}  contextual embeddings. In this study, we use the f1-score as we observed that the trends of recall, precision, and f1 are similar.

\end{itemize}
We calculate LENS through the authors' GitHub implementation\footnote{\url{https://github.com/Yao-Dou/LENS}} and BERTScore using the EASSE package~\cite{alva-manchego-etal-2019-easse}.

\subsubsection{Corpus-level Metrics}
\begin{itemize}
    \item \textbf{SARI}~\cite{xu-etal-2016-optimizing} evaluates a simplification model by comparing its outputs against the references and source sentences, focusing on the words that are added, kept, and deleted. Its values range from $0$ to $100$, with higher values indicating better quality.  
    \item \textbf{BLEU}~\cite{papineni-etal-2002-bleu} measures string similarity between references and outputs. Derived from the field of machine translation, it is designed to evaluate translation accuracy by comparing the match of n-grams between the candidate translations and reference translations. This metric has been employed to assess sentence simplification, treating the simplification process as a translation from complex to simple language. BLEU scores range from $0$ to $100$, with higher scores indicating better quality.
    \item \textbf{FKGL}~\cite{Kincaid1975Derivation} evaluates readability by combining sentence and word lengths. Lower values indicate higher readability. The FKGL score starts from \(-3.40\) and has no upper bound.
\end{itemize}
We utilize the EASSE package~\cite{alva-manchego-etal-2019-easse} to calculate these corpus-level metric scores.

\subsection{Sentence-Level Results}
To assess sentence-level metrics' ability on differentiating the sentence-level simplification quality, we explore the correlation between those metrics and human evaluations by employing the point-biserial correlation coefficient~\cite{glass1995statistical, linacre2008expected}, utilizing the scipy package~\cite{2020SciPy-NMeth} for calculation\footnote{The point-biserial correlation coefficient was chosen because our human labels are mostly binary while evaluation metric scores are continuous.}. This coefficient ranges from \(-1\) and \(+1\), where $0$ indicates no correlation. 

Specifically, our analysis aims to assess the efficacy of sentence-level metrics in three aspects:
\begin{enumerate}
\item Identification of the presence of errors.
\item Distinction between high-quality and low-quality simplification overall.
\item Distinction between high-quality and low-quality simplification within a specific dimension.
\end{enumerate}
Given the data imbalance between sentences with and without errors and between high-quality and low-quality simplification, we report our findings using both raw data and downsampled (DS) data to balance the number of class samples.

\subsubsection{Identification of the Presence of Errors}
For all $4,500$ simplification outputs in Task $1$, each simplification output is classified as containing errors (labeled as $1$) or no error (labeled as $0$). We then compute the correlation coefficients between these labels and the metric scores. The results, presented in Table~\ref{tab:metric_error}, indicate that none of the metrics effectively identify erroneous simplifications, as evidenced by point-biserial correlation coefficients being near zero.

% \begin{table}[t]
% \centering
% \caption{Point-biserial correlation between the presence of errors and sentence-level metrics scores, with downsampling (DS) numbers provided.}
% \label{tab:metric_error}
% \begin{tabular}{lcccccc}
% \toprule
% & \multicolumn{2}{c}{All} & \multicolumn{2}{c}{GPT-4} & \multicolumn{2}{c}{Control-T5} \\
% & Raw & DS ($511$) & Raw & DS ($182$) & Raw & DS ($329$) \\
% \midrule
% LENS & $-0.04$ & $-0.03$ & $-0.04$ & $-0.04$ & $-0.13$ & $-0.11$ \\
% BERT precision & $-0.07$ & $-0.11$ & $-0.10$ & $-0.16$ & $-0.05$ & $-0.03$ \\
% BERT recall & $-0.12$ & $-0.16$ & $-0.14$ & $-0.20$ & $-0.12$ & $-0.10$ \\
% BERT f1 & $-0.09$ & $-0.13$ & $-0.11$ & $-0.18$ & $-0.07$ & $-0.05$ \\
% \bottomrule
% \end{tabular}
% \end{table}

\begin{table}[t]
\centering
\caption{Point-biserial correlation between the presence of errors and sentence-level metrics scores, with downsampling (DS) numbers provided.}
\label{tab:metric_error}
\resizebox{\textwidth}{!}{%
\begin{tabular}{lcccccccccc}
\toprule
& \multicolumn{2}{c}{All} & \multicolumn{2}{c}{GPT-4} & \multicolumn{2}{c}{Qwen2.5-72B} & \multicolumn{2}{c}{Llama-3.2-3B} & \multicolumn{2}{c}{Control-T5} \\
& Raw & DS ($1072$) & Raw & DS ($240$) & Raw & DS ($181$) & Raw & DS ($332$) & Raw & DS ($319$) \\
\midrule
LENS & $-0.16$ & $-0.15$ & $-0.10$ & $-0.11$ & $-0.10$ & $-0.16$ & $-0.25$ & $-0.25$ & $-0.14$ & $-0.15$ \\
BERT f1 & $-0.12$ & $-0.13$ & $-0.12$ & $-0.08$ & $-0.03$ & $-0.09$ & $-0.20$ & $-0.22$ & $-0.12$ & $-0.11$ \\
\bottomrule
\end{tabular}%
}
\end{table}

\subsubsection{Distinction Between High-Quality and Low-Quality Simplifications Overall} 
\label{sec:destinction_overall}
We examine all $10,471$ model-generated simplification outputs in Task $2$. Each simplification output is classified as high quality (labeled as $1$) if it received a high rating (a score of $3$) from at least two out of three annotators across all dimensions (fluency, simplicity, and meaning preservation), and low quality (labeled as $0$) otherwise. 
We compute the correlation coefficients between these classifications and the metric scores. As we discussed in Section~\ref{sec:likert_scale_rating}, Newsela is different from other corpora in that it allows significant meaning loss to prioritize simplicity. To reduce bias stemming from this particular dataset, we also calculate the correlation after excluding Newsela, i.e., only on Turk, ASSET, and /or SimPA (denoted as `T\&A\&S' for overall and LLMs, and `T\&A' for Control-T5). The distributions of high-quality and low-quality outputs are generally highly imbalanced. For example, GPT-4 generates $2,593$ overall high-quality simplification outputs compared to just $299$ low-quality ones, potentially affecting correlation results. To address this, we only report correlations after downsampling.
For each evaluation, we further divide simplification outputs based on the model to determine if there are differences in the metrics' capabilities. The results are summarized in Table~\ref{tab:metric_rating}. 

\paragraph{Metrics Fail to Effectively Differentiate Between High- and Low-Quality}
LENS shows some ability to distinguish quality in Control-T5 simplification outputs ($0.34$ of correlation coefficient) but remains limited overall ($0.15$ in `All-DS', and $0.01$--$0.34$ across models). BERTScore generally performs better across all models, though its effectiveness is not reliable enough ($0.34$ in `All-DS', and $0.20$--$0.60$ across models). Both metrics exhibit slightly higher correlation when evaluating Control-T5 (LENS: $0.34$, BERT f1: $0.60$) and Llama-3.2-3B (LENS: $0.15$, BERT f1: $0.39$) compared to GPT-4 (LENS: $0.01$, BERT f1: $0.20$) and Qwen2.5-72B (LENS: $0.03$, BERT f1: $0.33$). However, their performance declines notably after removing Newsela-derived simplification outputs.

\begin{figure*}[t]
\centering
\begin{subfigure}[b]{0.49\textwidth} 
    \centering
    \includegraphics[width=\textwidth]{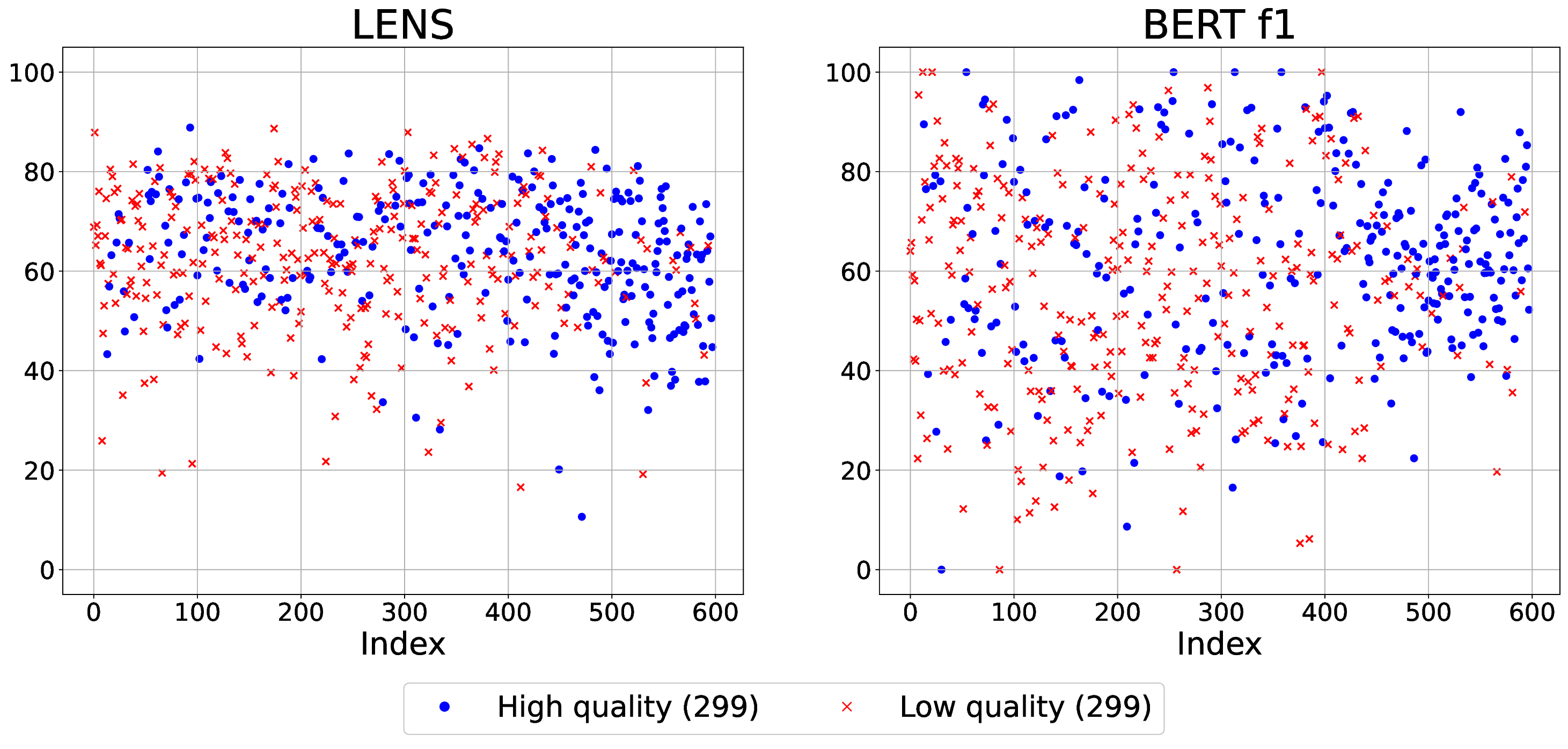} 
    \caption{GPT-4}
    \label{fig:meta-quality-gpt4}
\end{subfigure}
\hfill 
\begin{subfigure}[b]{0.49\textwidth} 
    \centering
    \includegraphics[width=\textwidth]{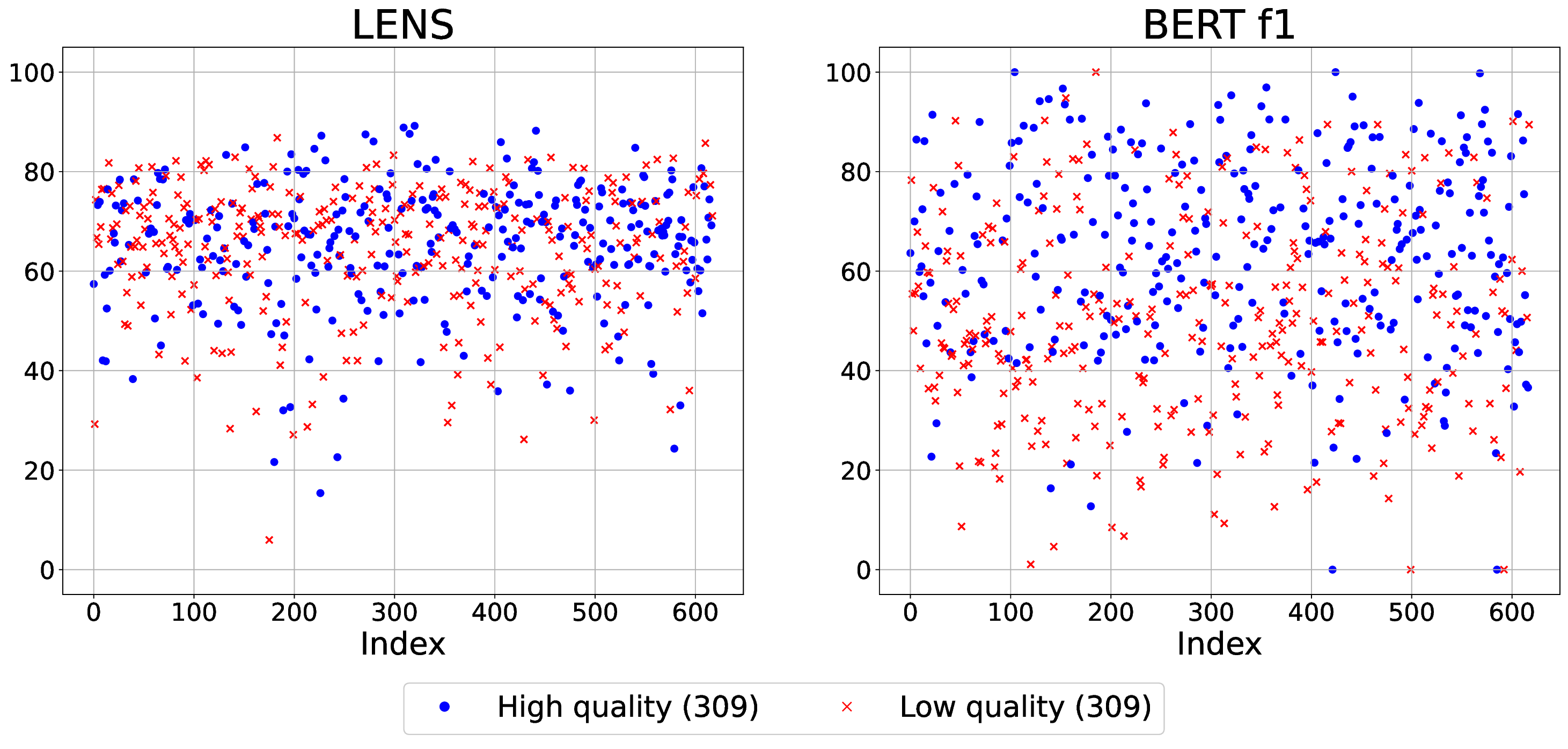} 
    \caption{Qwen2.5-72B}
    \label{fig:meta-quality-qwen}
\end{subfigure}
\hfill
\begin{subfigure}[b]{0.49\textwidth} 
    \centering
    \includegraphics[width=\textwidth]{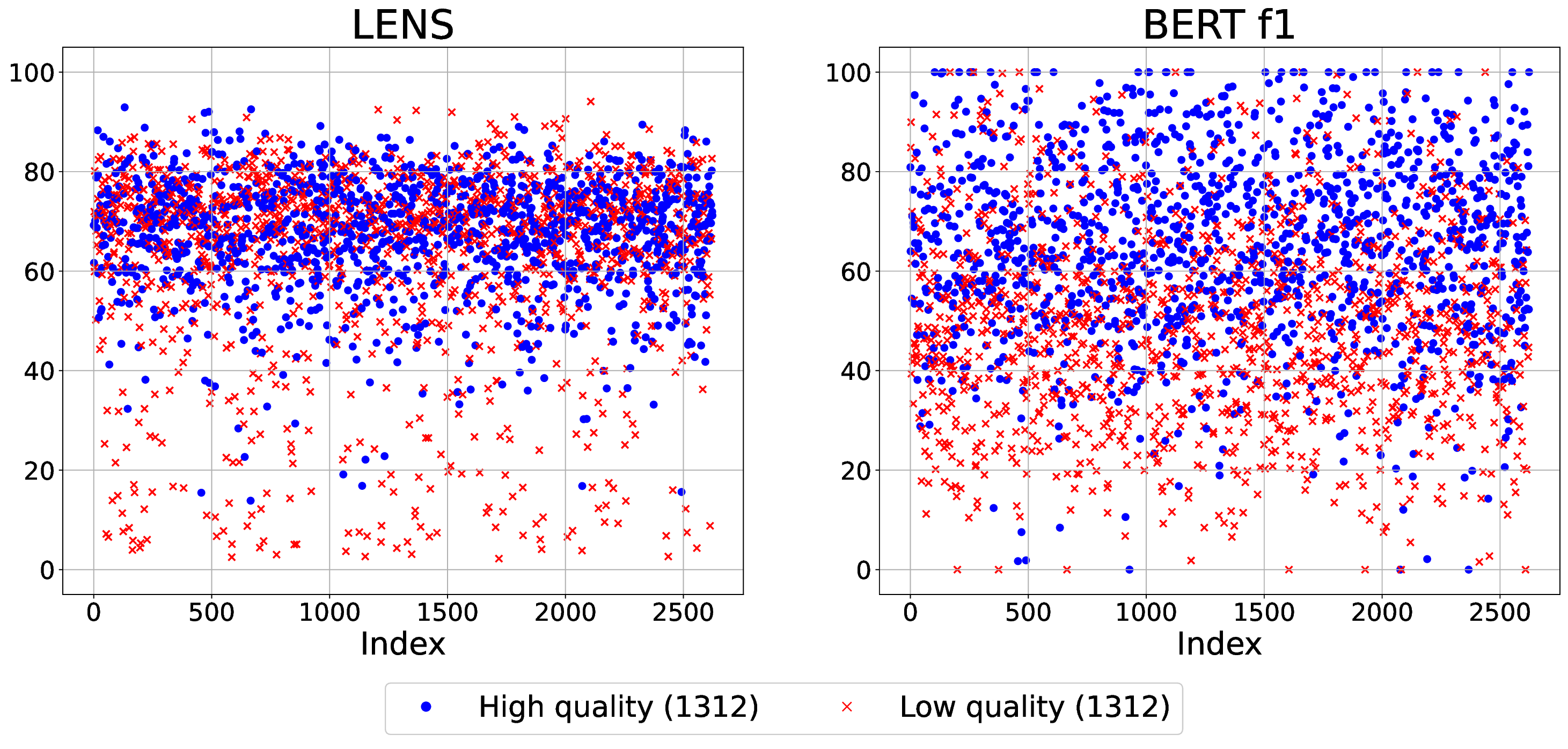} 
    \caption{Llama-3.2-3B}
    \label{fig:meta-quality-llama}
\end{subfigure}
\hfill
\begin{subfigure}[b]{0.49\textwidth} 
    \centering
    \includegraphics[width=\textwidth]{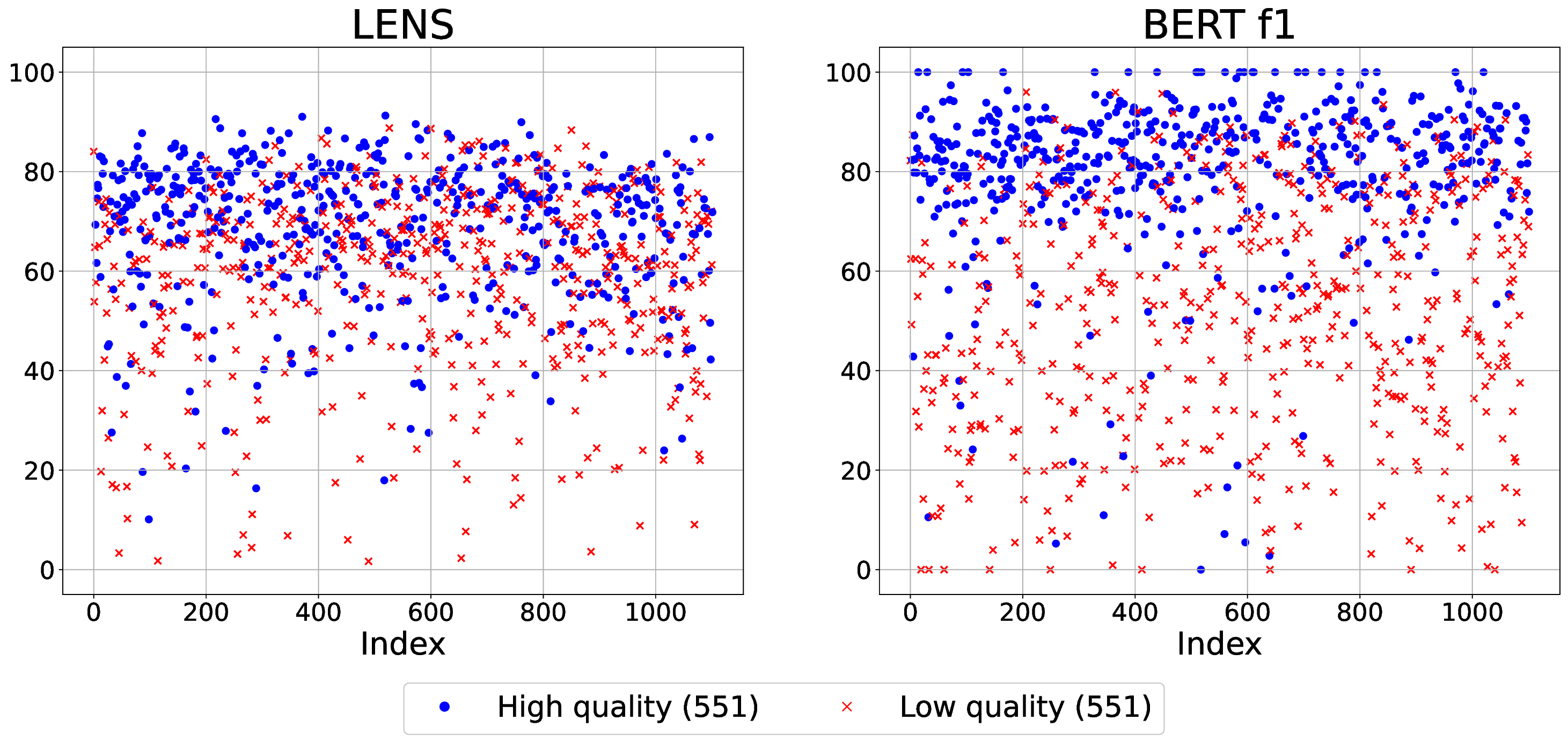} 
    \caption{Control-T5}
    \label{fig:meta-quality-t5}
\end{subfigure}
\caption{Comparative visualization of sentence-level metrics scores for high-quality (blue dot) vs. low-quality (red cross) simplification outputs}
\label{fig:meta-quality}
\end{figure*}

\paragraph{More Challenges in Evaluating High-quality Simplification Models}
To further compare the metrics' evaluations of high- and low-quality outputs across models, we incorporate visualizations (see Figure \ref{fig:meta-quality}) after downsampling. For GPT-4 and Qwen2.5-72B, regardless of the evaluation metric used, the scores of high and low-quality simplification outputs appear to blend, revealing a lack of discriminative capability. This indicates that these metrics struggle to differentiate when the overall quality is high, making them less suitable for evaluating advanced LLMs like GPT-4 and Qwen2.5-72B.
In contrast, for Llama-3.2-3B and Control-T5, high-quality sentence pairs rated by humans tend to receive higher scores from both metrics, with LENS showing a clearer alignment. However, low-quality sentence pairs rated by humans exhibit a wider range of scores, showing the metrics' limitations in capturing quality variations.

\begin{table}[t]
\centering
\caption{Point-biserial correlation between the overall human ratings of simplification outputs and sentence-level metrics scores}
\label{tab:metric_rating}
\resizebox{\textwidth}{!}{%
\begin{tabular}{lcccccccccc}
\toprule
& \multicolumn{2}{c}{All} & \multicolumn{2}{c}{GPT-4} & \multicolumn{2}{c}{Qwen2.5-70b} & \multicolumn{2}{c}{Llama-3.2-3B} & \multicolumn{2}{c}{Control-T5} \\
& \makecell{DS \\ (3,164)} & \makecell{T\&A\&S-DS \\ (936)} & \makecell{DS \\ (299)} & \makecell{T\&A\&S-DS \\ (104)} & \makecell{DS \\ (309)} & \makecell{T\&A\&S-DS \\ (72)} & \makecell{DS \\ (1,312)} & \makecell{T\&A\&S-DS \\ (530)} & \makecell{DS \\ (551)} & \makecell{T\&A-DS \\ (230)}\\
\midrule
LENS      & 0.15 & 0.01 & 0.01 & 0.06 & 0.03 & 0.18 & 0.15 & -0.07 & 0.34 & 0.18 \\
BERT f1   & 0.34 & 0.16 & 0.20 & -0.07 & 0.33 & 0.21 & 0.39 & 0.27  & 0.60 & 0.37 \\
\bottomrule
\end{tabular}%
}
\end{table}

\subsubsection{Distinction Between High-Quality and Low-Quality Simplifications Within a Specific Dimension} 
We examined $10,471$ model-generated simplification outputs in Task $2$ across individual dimensions. For each dimension, simplification outputs are classified as high quality (labeled as $1$) if they received a high rating (a score of $3$) from at least two out of three annotators, and low quality (labeled as $0$) otherwise. 
Based on our classification, on fluency, only one GPT4, one Qwen2.5-70b, two Llama-3.2-3B, and five Control-T5-generated simplification outputs are low quality. Given that these models rarely generate disfluent outputs, we focus on the dimensions of meaning preservation and simplicity. We then compute the correlation between these ratings and metrics scores.
As in Section~\ref{sec:destinction_overall}, we only report results after downsampling and incorporate results after excluding Newsela.

\begin{table}[t]
\small
\centering
\caption{Point-biserial correlation between sentence-level scores and human ratings in single dimension}
\begin{subtable}{\textwidth}
\small
\centering
\caption{Meaning preservation}
\label{tab:metric_rating_mp}
\resizebox{\textwidth}{!}{%
\begin{tabular}{lcccccccccc}
\toprule
& \multicolumn{2}{c}{All} & \multicolumn{2}{c}{GPT-4} & \multicolumn{2}{c}{Qwen2.5-72B} & \multicolumn{2}{c}{Llama-3.2-3B} & \multicolumn{2}{c}{Control-T5} \\
& \makecell{DS \\ ($2,848$)} & \makecell{T\&A\&S-DS \\ ($767$)} & \makecell{DS \\ ($168$)} & \makecell{T\&A\&S-DS \\ ($37$)} & \makecell{DS \\ ($263$)} & \makecell{T\&A\&S-DS \\ ($43$)} & \makecell{DS \\ ($1,187$)} & \makecell{T\&A\&S-DS \\ ($471$)} & \makecell{DS \\ ($565$)} & \makecell{T\&A-DS \\ ($216$)}\\
\midrule
LENS      & $0.11$ & $-0.09$ & $-0.04$ & $-0.02$ & $-0.02$ & $0.10$ & $0.06$ & $-0.06$ & $0.31$ & $0.19$ \\
BERT f1   & $0.37$ & $0.24$  & $0.40$  & $0.30$  & $0.48$  & $0.33$ & $0.42$ & $0.34$  & $0.58$ & $0.38$ \\
\bottomrule
\end{tabular}%
}

\end{subtable}

\hfill 

\begin{subtable}{\textwidth}
\small
\centering
\caption{Simplicity}
\label{tab:metric_rating_simp}
\resizebox{\textwidth}{!}{%
\begin{tabular}{lcccccccccc}
\toprule
& \multicolumn{2}{c}{All} & \multicolumn{2}{c}{GPT-4} & \multicolumn{2}{c}{Qwen2.5-72B} & \multicolumn{2}{c}{Llama-3.2-3B} & \multicolumn{2}{c}{Control-T5} \\
& \makecell{DS \\ ($451$)} & \makecell{T\&A\&S-DS \\ ($188$)} & \makecell{DS \\ ($135$)} & \makecell{T\&A\&S-DS \\ ($68$)} & \makecell{DS \\ ($47$)} & \makecell{T\&A\&S-DS \\ ($30$)} & \makecell{DS \\ ($242$)} & \makecell{T\&A\&S-DS \\ ($65$)} & \makecell{DS \\ ($27$)} & \makecell{T\&A-DS \\ ($25$)}\\
\midrule
LENS    & $0.43$ & $0.28$ & $0.19$ & $0.21$ & $0.43$ & $0.48$ & $0.57$ & $0.23$ & $0.16$ & $0.31$ \\
% BERT precision & $0.19$ & $0.15$ & $0.13$ & $0.16$ & $0.13$ & $0.32$ & $0.42$ & $-0.01$ & $-0.14$ & $0.14$ \\
% BERT recall & $-0.0$ & $-0.14$ & $0.02$ & $-0.04$ & $-0.11$ & $-0.08$ & $0.07$ & $-0.34$ & $-0.24$ & $0.15$ \\
BERT f1 & $0.06$ & $-0.29$ & $-0.04$ & $-0.33$ & $-0.02$ & $-0.06$ & $0.26$ & $-0.39$ & $-0.19$ & $0.50$ \\
\bottomrule
\end{tabular}%
}
\end{subtable}
\end{table}

Table~\ref{tab:metric_rating_mp} indicates results for meaning preservation. 
Overall, while moderate, BERTScore shows a stronger correlation with human evaluations of meaning preservation ($0.37$ in `All-DS' and $0.40$--$0.58$ across models) compared to LENS scores ($0.11$ in `All-DS' and $-0.04$--$0.31$ across models). However, these correlations decline from $0.37$--$0.58$ to $0.24$–-$0.38$ when simplification outputs from Newsela are excluded. Table~\ref{tab:metric_rating_simp} shows results for simplicity. Overall, LENS demonstrates a stronger correlation with human evaluations for simplicity ($0.43$ in `All-DS', and $0.16$--$0.57$ across models) compared to BERT score ($0.06$ in `All-DS', and $-0.19$--$0.26$ across models ).

\subsection{Corpus-level Results} 
\label{subsec:autoresult}
Our human evaluations reveal GPT-4 and Qwen2.5-72B's simplification outputs are generally superior, evidenced by fewer errors, better meaning preservation, and comparable fluency and simplicity to those generated by Llama-3.2-3B and Control-T5. 
In this section, we compare the corpus-level metrics scores of the models with human evaluation results to determine if they align, i.e., whether they rate GPT-4 and Qwen2.5-72B higher than Llama-3.2-3B and Control-T5.

\begin{table}[t]
\parbox[t]{.45\linewidth}{
\small
\centering
\caption{Corpus-level scores for different models}
\label{tab:system_performance}
\begin{tabular}{lllll}
\toprule
\textbf{} & \textbf{Model} & \textbf{SARI} & \textbf{BLEU} & \textbf{FKGL} \\
\midrule
\multicolumn{1}{l}{Turk} &
GPT-4 & $42.9$ & $\mathbf{72.0}$ & $8.3$ \\
\multicolumn{1}{l}{} &
Qwen2.5-72B & $42.7$ & $63.6$ & $7.8$ \\
\multicolumn{1}{l}{} &
Llama-3.2-3B & $38.2$ & $55.8$ & $7.5$ \\
\multicolumn{1}{l}{} &
Control-T5 & $\mathbf{43.7}$ & $68.2$ & $\mathbf{5.8}$ \\

\midrule
\multicolumn{1}{l}{ASSET} &
GPT-4 & $47.3$ & $59.3$ & $7.6$ \\
\multicolumn{1}{l}{} &
Qwen2.5-72B & $\mathbf{47.9}$ & $69.8$ & $8.2$ \\
\multicolumn{1}{l}{} &
Llama-3.2-3B & $45.5$ & $67.8$ & $8.2$ \\
\multicolumn{1}{l}{} &
Control-T5 & $44.9$ & $\mathbf{74.5}$ & $\mathbf{6.3}$ \\

\midrule
\multicolumn{1}{l}{Newsela} &
GPT-4 & $41.4$ & $13.9$ & $5.7$ \\
\multicolumn{1}{l}{} &
Qwen2.5-72B & $41.7$ & $17.0$ & $6.3$ \\
\multicolumn{1}{l}{} &
Llama-3.2-3B & $\mathbf{41.9}$ & $13.9$ & $\mathbf{4.1}$ \\
\multicolumn{1}{l}{} &
Control-T5 & $38.6$ & $\mathbf{24.0}$ & $4.2$ \\

\midrule
\multicolumn{1}{l}{SimPA} &
GPT-4 & $40.8$ & $23.5$ & $9.6$ \\
\multicolumn{1}{l}{} &
Qwen2.5-72B & $\mathbf{42.8}$ & $\mathbf{28.0}$ & $10.3$ \\
\multicolumn{1}{l}{} &
Llama-3.2-3B & $38.2$ & $18.9$ & $\mathbf{8.5}$ \\

\bottomrule
\end{tabular}
}
\hfill
\parbox[t]{.45\linewidth}{
\small
\centering
\caption{Comparison of sentence simplification models and LLMs on the Turk corpus. The first section presents SARI scores measured by Alva-Manchego et al.~\cite{alva-manchego-etal-2020-data}, while the second section shows SARI scores for LLMs computed in our study.}
\label{tab:compare_ss_sytems}
\begin{tabular}{lccl}\toprule
\textbf{System}  & \textbf{SARI}  \\ 
\midrule
Hybrid~\cite{narayan-gardent-2014-hybrid}                   & $31.4$                                         \\ 
DRESS-LS~\cite{zhang-lapata-2017-sentence}                  & $37.3$                                         \\ 
PBSMT-R~\cite{wubben-etal-2012-sentence}                     & $38.6$                                     \\ 
SBSMT-SARI~\cite{xu-etal-2016-optimizing}                  & $40.0$                                      \\ 
DMASS-DCSS~\cite{zhao-etal-2018-integrating}                  & $40.4$                                        \\ 
\midrule
Llama-3.2-3B           & $38.2$                                        \\ 
Qwen2.5-72B            & $42.7$                                        \\ 
GPT-4                  & $42.9$                                        \\ 

\bottomrule
\end{tabular}
}
\end{table}

The metrics' scores are detailed in Table~\ref{tab:system_performance}, with the best scores emphasized in bold. 
\paragraph{SARI Shows Effectiveness on Transformation-Rich Datasets}
\textbf{SARI} favors GPT-4 and Qwen2.5-72B over Llama-3.2-3B and Control-T5 in transformation-rich datasets --- ASSET and SimPA, aligning with our human evaluations of overall superior performance. 
In Turk, which focuses on lexical paraphrasing, GPT-4, Qwen2.5-72B, and Control-T5 are scored similarly, all higher than Llama-3.2-3B.
% Although overall GPT-4 and Qwen2.5-72B surpass Control-T5 in Turk, our error analysis points out that a significant portion of the errors in these two models' outputs stem from the use of more complex lexical expressions. 
Considering that our error analysis points out that GPT-4 and Qwen2.5-72B tend to use more complex lexical expressions, SARI may preserve sensitivity on lexical simplicity, the same as what was reported in a previous study~\cite{xu-etal-2016-optimizing}. In Newsela, GPT-4, Qwen2.5-72B and Llama-3.2-3B are scored similarly, all higher than Control-T5. Although overall GPT-4 and Qwen2.5-72B surpass Llama-3.2-3B in Newsela, our error analysis finds that sometimes Llama-3.2-3B merges multiple simplifications into a single output, resulting in repetitive content (see Section \ref{sec:errors_models}). This behavior may inflate Llama-3.2-3B's SARI score on Newsela. Before the emergence of LLMs, sentence simplification learned transformation rules from parallel corpora of original-simplified sentence pairs~\cite{alva-manchego-etal-2020-data}. We compare LLM-generated simplifications with those from well-established models by examining SARI scores reported by Alva-Manchego et al.~\cite{alva-manchego-etal-2019-easse} on the Turk Corpus. As shown in Table~\ref{tab:compare_ss_sytems}, larger LLMs such as Qwen2.5-72B and GPT-4 surpass all previously reported systems in terms of SARI scores, while the smaller Llama-3.2-3B model demonstrates comparable performance. These findings are consistent with our human evaluation results.

\paragraph{BLEU is Unsuitable}
\textbf{BLEU} significantly favors Control-T5 on ASSET and Newsela, which does not match our human evaluations. Studies~\cite{xu-etal-2016-optimizing, sulem-etal-2018-bleu} have demonstrated that BLEU is unsuitable for simplification tasks, as it tends to negatively correlate with simplicity, often penalizing simpler sentences, and gives high scores to sentences that are close or even identical to the input. Our finding further underscores the limitations of BLEU in evaluating sentence simplification. 
\paragraph{Limitations of FKGL in Comprehensive Quality Evaluation}
\textbf{FKGL} ranks Control-T5's outputs as easier to read compared to those from GPT-4 and Qwen2.5-72B across all datasets. This aligns with our human evaluation; Control-T5 tends to generate simpler sentences at the expense of meaning preservation, thereby making the sentence easier to read. However, FKGL's focus solely on readability, without taking into account the quality of the content or the reference sentences, limits its effectiveness in a comprehensive quality analysis. Previous studies~\cite{alva-manchego-etal-2021-un, tanprasert-kauchak-2021-flesch} show that FKGL is unsuitable for sentence simplification evaluation. Our finding further highlights its limitations in accurately evaluating corpus-level sentence simplification.

\subsection{Summary of Findings} 
We summarize our findings on the meta-evaluation of existing evaluation metrics for sentence simplification, namely LENS, BERTScore, SARI, BLEU, and FKGL.
\begin{enumerate}
    \item Existing metrics are not capable of identifying the presence of errors in sentences. 
    \item At the overall sentence level, both LENS and BERTScore fail to effectively differentiate between high-quality and low-quality simplifications, particularly when evaluating high-quality simplification models, i.e., GPT-4 and Qwen2.5-72B.
    \item For sentence-level meaning preservation, BERTScore is better at distinguishing high-quality from low-quality simplifications compared to LENS. Conversely, for sentence-level simplicity, LENS is more effective than BERTScore. However, the correlations of both metrics against human ratings are limited to the moderate range.
    \item At the corpus level, SARI aligns more closely with our human evaluations, while BLEU appears less suitable.
\end{enumerate}

\section{Conclusion}
In this study, we conduct an in-depth human evaluation of LLMs in sentence simplification. Our findings highlight that LLMs surpass the previous SOTA model, Control-T5, by generating fewer erroneous simplification outputs and preserving the source sentence's meaning better. These results underscore the superiority of advanced LLMs in this task. Among LLMs, while larger LLMs generally perform better, performance does not always scale directly with size. Medium-sized LLMs may offer strong potential in simplification tasks. Nevertheless, we observed limitations in large and medium-sized LLMs, notably in GPT-4 and Qwen2.5-72B's handling of lexical paraphrasing. Further, our meta-evaluation of sentence simplification's automatic metrics demonstrates their inadequacy in accurately assessing the quality of LLM-generated simplifications.

With their advanced capabilities, LLMs hold great promise as tools for text simplification, benefiting non-native speakers and individuals with reading difficulties. However, the limitations identified in our study underscore the need for careful selection and application of these models to ensure they genuinely benefit users. For instance, models like Qwen2.5-72B and GPT-4 may be preferable to others. Even with these advanced models, caution is necessary to mitigate paraphrasing errors, as making lexical expressions more complex is counterproductive. We hope our study provides valuable insights for future research, contributing to improved simplification performance in LLMs and enhancing their usability and effectiveness across diverse user groups. Our investigation opens up multiple directions for future research. Future studies could investigate how to mitigate lexical paraphrasing issues. For example, it would be worthwhile to explore whether fine-tuning could help. Moreover, there's a need for more sensitive automatic metrics to evaluate the sentence-level quality of simplifications generated by LLMs properly. Automating our error classification approach could enable real-time monitoring, allowing a plugin to periodically sample model-generated simplifications, identify error types and locations, and incorporate them into instructions to help the model self-correct.

\begin{acks}
This work was supported by JSPS KAKENHI Grant Number JP21H03564 and JP25K03233.
\end{acks}

\bibliographystyle{ACM-Reference-Format}
\bibliography{ref}

\appendix

\section{Details of Models}
\subsection{Best Prompts in GPT-4's prompt engineering}
\label{appendix:prompts}
\begin{figure}[t]
    \centering
    \parbox[t]{0.3\textwidth}{%
        \centering
        \includegraphics[width=\linewidth]{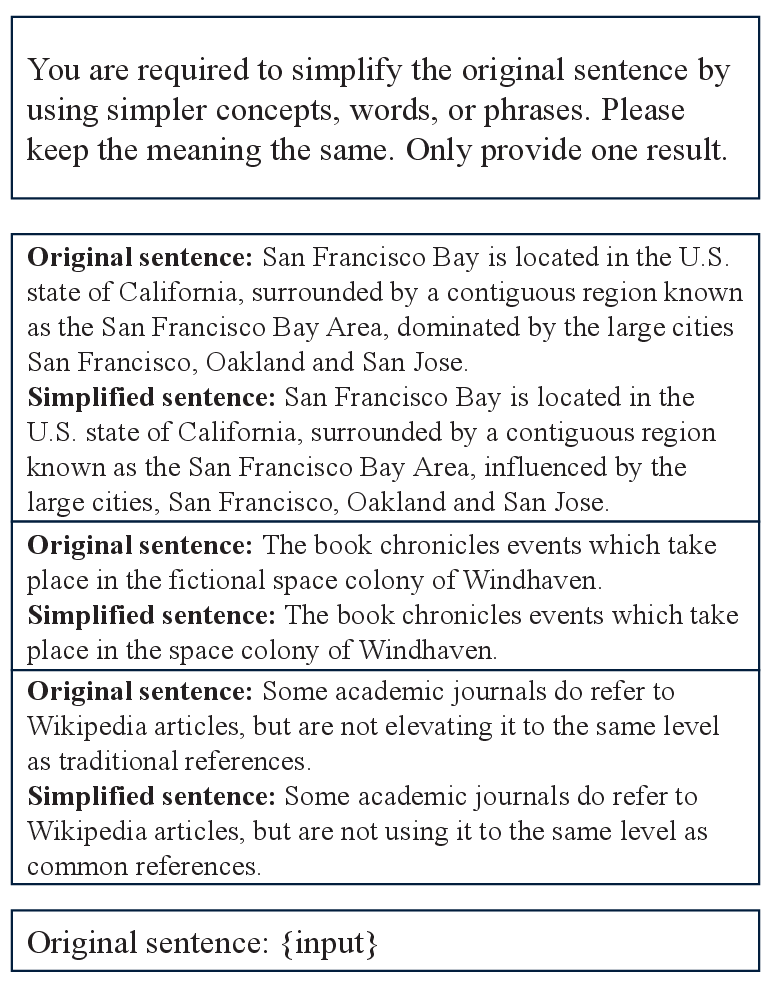}
        \caption{Prompt for Turk}
        \label{fig:turk-prompt}
    }
    \parbox[t]{0.3\textwidth}{%
        \centering
        \includegraphics[width=\linewidth]{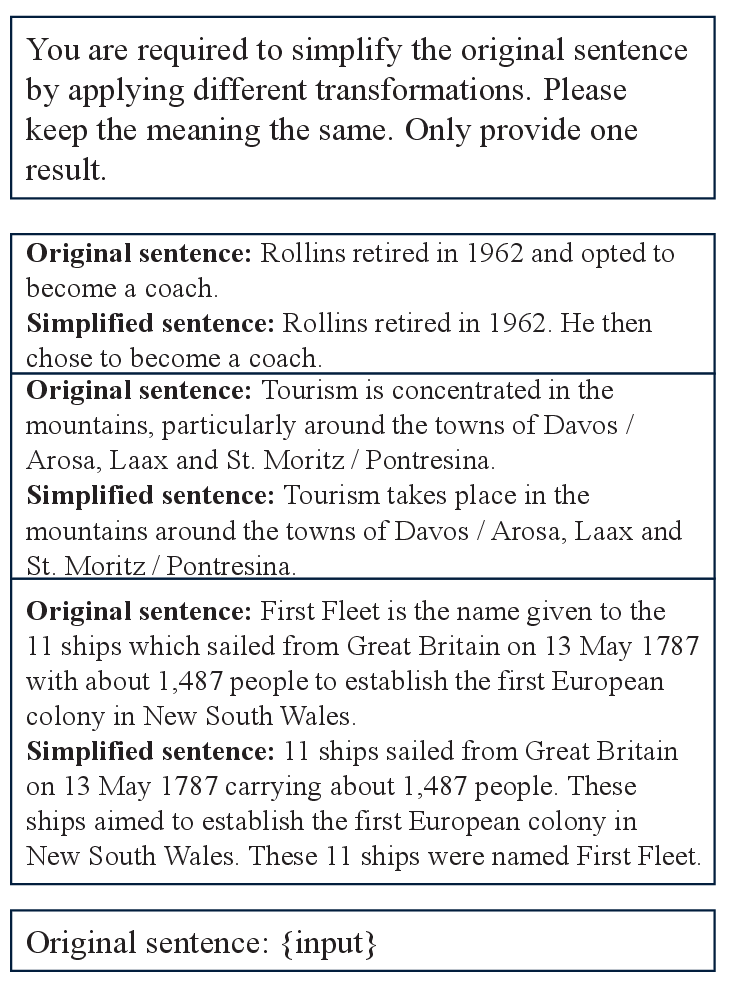}
        \caption{Prompt for ASSET}
        \label{fig:asset-prompt}
    }
    \parbox[t]{0.35\textwidth}{%
        \centering
        \includegraphics[width=\linewidth]{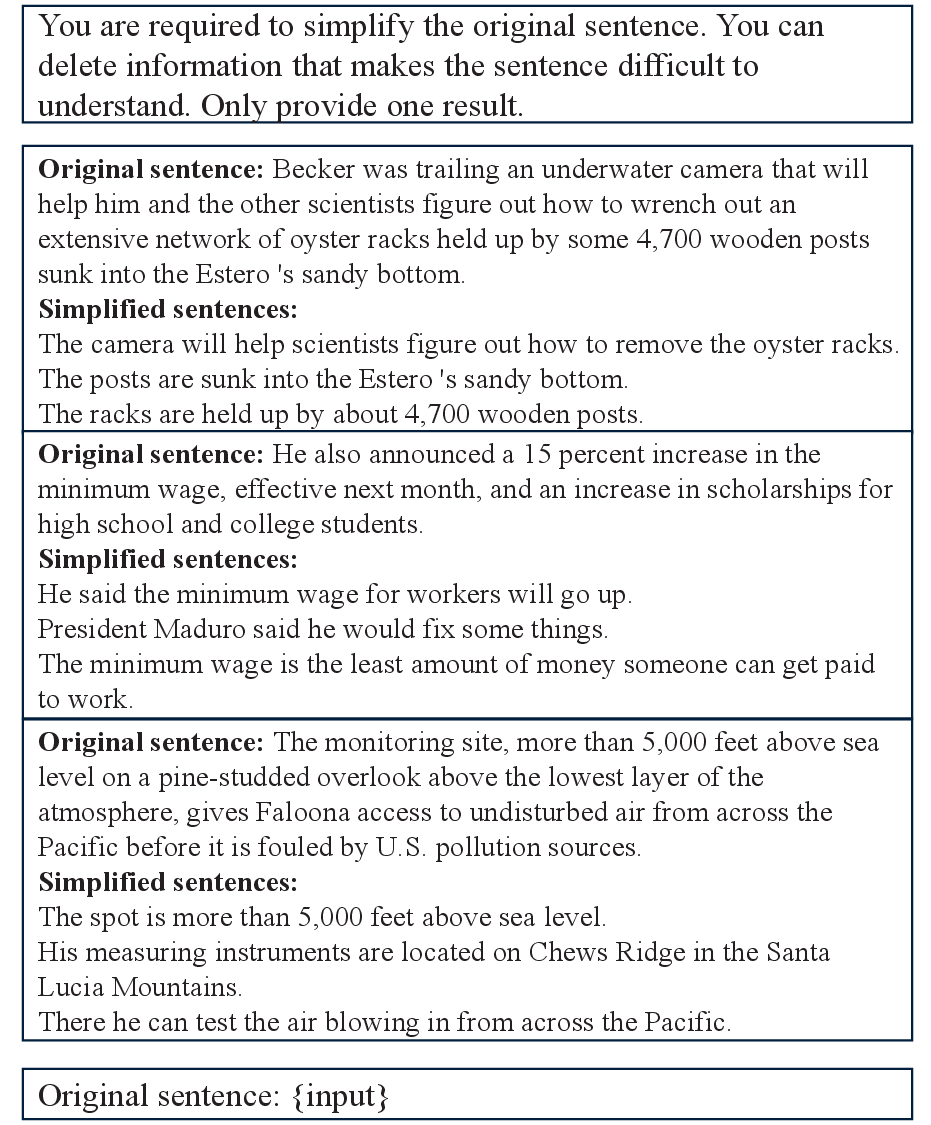}
        \caption{Prompt for Newsela}
        \label{fig:newsela-prompt}
    }
\end{figure}

Figure~\ref{fig:turk-prompt}, ~\ref{fig:asset-prompt}, and ~\ref{fig:newsela-prompt} illustrate the best prompts that achieved the highest SARI scores on each validation set during GPT-4's prompt engineering. Each prompt comprises: instructions, examples of original to simplification(s) transformation, and a source sentence.

\subsection{Optimal Configuration of Replicated Control-T5}
\label{appendix:t5_config}
Control-T5 was trained on WikiLarge in original implementation~\cite{sheang-saggion-2021-controllable}. While we followed the methodology described in the original paper, we made a few modifications. Specifically, we adjusted the learning rate to $1e-4$ and set the batch size to $16$, which brought our results more in line with those reported in the original study.
We further incorporated Newsela. The optimal configuration consists of a batch size of $16$, training over $16$ epochs, and a learning rate of $2.16e-05$. Additionally, the specific control token ratios are as follows: CharRatio at $0.4$, LevenshteinRatio at $0.7$, WordRankRatio at $1.35$, and DepthTreeRatio at $1.25$.

\section{Annotation Guidelines for Task 2}
\label{appendix:annotation}
\begin{figure}[t]
\centering
\includegraphics[width=\linewidth]{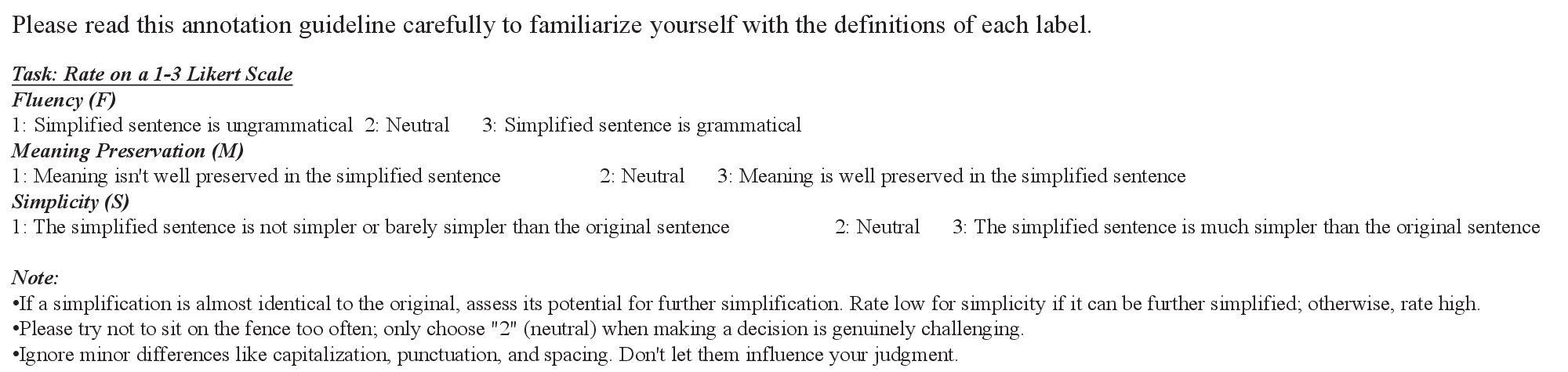}
\caption{Annotation guidelines in Task $2$}
\label{fig:task2_guidelines}
\end{figure}

% This section provides guidelines provided to annotators in Task $2$. The guidelines provided in Task $1$ are the definition and examples of each error type detailed in Table~\ref{tab:definition_errors}. 
Figure~\ref{fig:task2_guidelines} shows the guidelines provided to annotators in Task $2$. We also provided the same annotation examples, including source-simple pairs, ratings, and explanations, as those in~\cite{kriz-etal-2019-complexity, jiang2020neural}.

\end{document}